\documentclass[10pt, a4paper]{article}

\usepackage{lrec2026}

\usepackage[utf8]{inputenc}
\usepackage{booktabs}
\usepackage{relsize}
\usepackage{xcolor}
\usepackage{lipsum}
\usepackage{todonotes}
\usepackage{adjustbox}
\usepackage{multirow}
\usepackage{color, colortbl}
\usepackage{adjustbox}

\newcommand{\HPLT}{HPLT~3.0}

\newcommand{\sref}[1]{\S$\,$\ref{#1}}

\title{HPLT~3.0:\ Very Large-Scale Multilingual Resources for LLMs and MT\\[0.5ex]
\smaller Mono- and Bi-lingual Data, Multilingual Evaluation, and Pre-Trained Models}

\name{\smaller
  Stephan Oepen$^{\clubsuit}$,
  Nikolay Arefev$^{\clubsuit}$,
  Mikko Aulamo$^{\spadesuit}$,
  Marta Bañón$^{\heartsuit}$,
  Maja Buljan$^{\clubsuit}$,\\ \bfseries
  Laurie Burchell$^{\diamondsuit}$,
  Lucas Charpentier$^{\clubsuit}$,
  Pinzhen Chen$^{\bullet}$,
  Mariia Fedorova$^{\clubsuit}$,
  Ona de Gibert$^{\spadesuit}$,\\ \bfseries
  Barry Haddow$^{\bullet}$,
  Jan Haji\v{c}$^{\circ}$,
  Jind\v{r}ich Helcl$^{\clubsuit}$,
  Andrey Kutuzov$^{\clubsuit}$,
  Veronika Laippala$^{\star}$,
  Zihao Li$^{\spadesuit}$,\\ \bfseries
  Risto Luukkonen$^{\star}$,
  Bhavitvya Malik$^{\bullet}$,
  Vladislav Mikhailov$^{\clubsuit}$,
  Amanda Myntti$^{\star}$,\\ \bfseries
  Dayyán O'Brien$^{\bullet}$,
  Lucie Poláková$^{\circ}$,
  Sampo Pyysalo$^{\star}$,
  Gema Ram{\'i}rez S{\'a}nchez$^{\heartsuit}$,\\ \bfseries
  Janine Siewert$^{\spadesuit}$,
  Pavel Stepachev$^{\bullet}$,
  Jörg Tiedemann$^{\spadesuit}$,
  Teemu Vahtola$^{\spadesuit}$,\\ \bfseries
  Du\v{s}an Vari\v{s}$^{\circ}$,
  Fedor Vitiugin$^{\star}$,
  Tea Vojtěchová$^{\circ}$,
  Jaume Zaragoza$^{\heartsuit}$\\[0.5ex]}

\address{\mbox{}$^{\clubsuit}$ University of Oslo, Department of Informatics\\
  \mbox{}$^{\spadesuit}$ University of Helsinki, Department of Digital Humanities\\
  \mbox{}$^{\heartsuit}$ Prompsit Language Engineering\\
  \mbox{}$^{\diamondsuit}$ The Common Crawl Foundation\\
  \mbox{}$^{\bullet}$ Edinburgh University, School of Informatics\\
  \mbox{}$^{\circ}$ Charles University, Prague, Institute of Formal and Applied Linguistics\\
  \mbox{}$^{\star}$ TurkuNLP, University of Turku, Department of Computing\\
         \texttt{oe@ifi.uio.no}}

\abstract{We present an ongoing initiative to provide open, very large,
  high-quality, and richly annotated textual datasets for almost 200 languages.
  At 30 trillion tokens, this is likely the largest generally available
  multilingual collection of LLM pre-training data.
  These datasets are derived from web crawls from different sources and
  accompanied with a complete, open-source pipeline for document selection from
  web archives, text extraction from HTML, language identification for noisy
  texts, exact and near-deduplication, annotation with, among others, register
  labels, text quality estimates, and personally identifiable information; and
  final selection and filtering.
  We report on data quality probes through contrastive and analytical
  statistics, through manual inspection of samples for 24 languages, and
  through end-to-end evaluation of various language model architectures trained
  on this data.
  For multilingual LLM evaluation, we provide a comprehensive collection of
  benchmarks for nine European languages, with special emphasis on natively
  created tasks, mechanisms to mitigate prompt sensitivity, and refined
  normalization and aggregation of scores.
  Additionally, we train and evaluate a family of near 60 monolingual
  encoder and encoder–decoder models, as well as a handful of monolingual GPT-like
  reference models.
  Besides the monolingual data and models, we also present a very large
  collection of parallel texts automatically mined from this data, together
  with a novel parallel corpus synthesized via machine translation.\\ \newline
  }

\begin{document}

\maketitleabstract

\section{Introduction} 
\label{sc:introduction}

Massive text collections for pre-training are the ``crude oil'' of the LLM era.
The process of ``refining'' high-quality datasets from web data at
scale presupposes computational infrastructure and technological muscle that
oftentimes is characteristic of corporate involvement, as evidenced for example by some
notable generally available pre-training datasets:\ C4 (the Colossal Clean
Crawled Corpus, by Google and Allen AI; \citealp{JMLR:v21:20-074}), FineWeb~1
\& 2 (by Hugging Face;
\citealp{penedo2024fineweb,penedo2025fineweb2pipelinescale}),
MADLAD-400 (by Google; \citealp{10.5555/3666122.3669062}), or
Nemotron-CC (by Nvidia; \citealp{su-etal-2025-nemotron}).
With notable exceptions, this line of work tends to capitalize on the
English language.

We build on the open-source pipeline of the European academic consortium HPLT
(High-Performance Language Technologies;
\citealp{de-gibert-etal-2024-new,burchell-etal-2025-expanded}), a project
that was funded under the Horizon Europe programme in 2022--2025.
We revise and improve various of the sub-components, and apply the updated
pipeline to a greatly enlarged collection of raw web data.
This results in a massive pre-training dataset of high-quality texts in close
to 200 distinct language--script combinations.
We dub this novel resource \HPLT.
The data comprises some 30 trillion sub-word tokens in total, of which close to
half represent languages other than English.


Following the examples of FineWeb and HPLT~2.0, we make this resource publicly
available under the most permissive terms of use possible (see \sref{sc:ethics}
below).\footnote{\mbox{\url{https://hplt-project.org/datasets/v3.0}}} 
We further share the refined open-source processing pipeline, a novel
multilingual evaluation framework, as well as various collections of language
models pre-trained on \HPLT\ data.
Finally, while our focus in this report is on the monolingual data and models,
we also briefly summarize ongoing work to derive novel bilingual datasets for
28 language pairs, provide associated machine translation models,
and synthesize additional pre-training data for underrepresented languages by
machine translation of very high-quality English documents.
In our view, it is the totality of generally available and very large-scale
resources and the documentation of the underlying processes that bears promise
of ``democratizing'' the current LLM and MT landscape.
In-depth technical and experimental details are provided through the appendices.

\section{Raw Web Archive Data} 
\label{sc:sources}

There are few available collections of massive web archives.
Our work builds on the same set of so-called ``wide crawls'' from the Internet
Archive (IA) as the HPLT~2.0 release, but combines this with a broader and much
larger set of snapshots from the Common Crawl (CC).
Specifically, we start from some 3.3 petabytes of IA crawls from the
period 2012--2020.
We complement this data with 57 CC full snapshots from the period 2014--2025,
making sure to include all available data since 2020, together with about half
of the earlier CC snapshots.
This amounts to almost five times more raw CC data than in HPLT~2.0, bringing our
total volume of web archives to about 7.2 petabytes.
\HPLT\ to date does not include a few experimental data sources, an
IA ``survey crawl'' and a minor sample from the ArchiveBot repository (270
terabytes in HPLT).
We plan on reporting on an enlarged dataset in 2026, augmented with
another 3 petabytes from ArchiveBot.

\section{Monolingual Data Preparation}
\label{sc:processing}

\begin{figure}
    \centering
    \includegraphics[width=\linewidth]{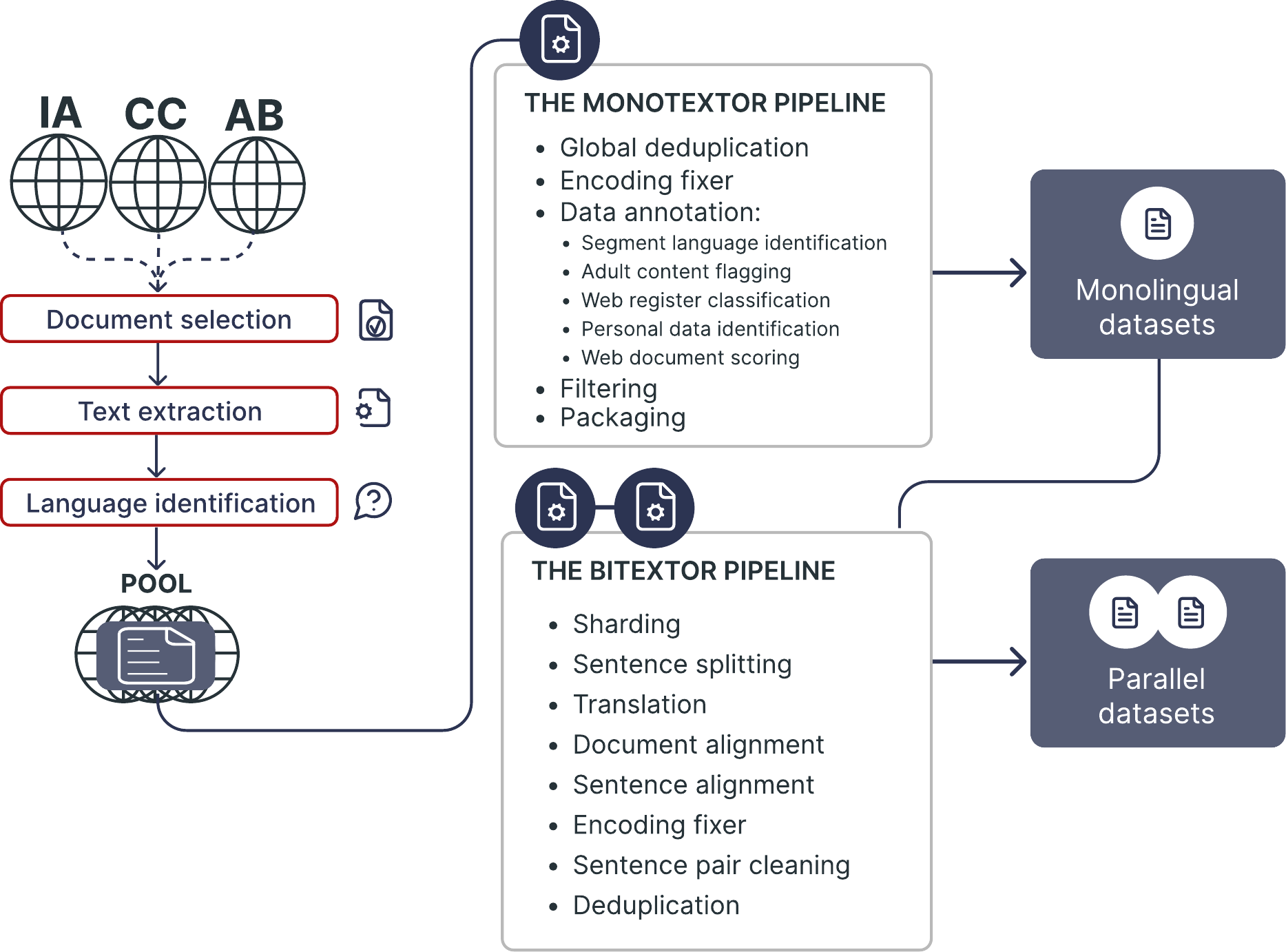}
    \caption{Schematic overview of data preparation.}
    \label{fg:pipeline}
\end{figure}

Extraction of high-quality and richly annotated text from raw web archives
proceeds through a sequence of refinement and filtering steps.\footnote{\url{https://github.com/hplt-project/HPLT-textpipes}}
Figure~\ref{fg:pipeline} shows the main components of our data preparation pipeline, 
which is an updated and extended version of the one used in HPLT~2.0 \cite{burchell-etal-2025-expanded}.
The following paragraphs provide a high-level overview of the refinements we
have made to the original HPLT~2.0 pipeline.

\paragraph{Text Extraction} 
%
%

Like FineWeb and HPLT~2.0, we stick to the Trafilatura text extraction framework
\cite{barbaresi-2021-trafilatura}, but we conduct a comprehensive data-driven
optimization of its many hyper-parameters, emphasizing extraction quality over
speed and text recall.

\paragraph{Language Identification} 

We predict the language of the text during preprocessing using OpenLID-v2, an updated version of the model described in \citet{burchell-etal-2023-open}. Compared to the prior version of OpenLID, we revise the inventory of language labels to align with the Flores$^+$ evaluation set \cite{burchell-etal-2024-findings}, and include additional training data for Dari, South Azerbaijani, Asturian, and Paraguayan Guaran\'{i}. We also revise input preprocessing for language identification to a simpler, language-agnostic pipeline: We normalize whitespace, apply lowercasing, and remove non-word characters and digits. This change increases the robustness of language identification to the non-standard orthography commonly found in web documents. 

\paragraph{Deduplication} 

Random repetitions caused by overlapping crawls can negatively impact LLM
training \cite{lee-etal-2022-deduplicating}.
While HPLT~2.0 followed the original FineWeb design and limited itself to
per-collection or per-crawl deduplication, we implement MinHash-based global 
near-deduplication for all languages in \HPLT\ except English, Russian,
and Chinese for which we stick to per-crawl deduplication for computational
efficiency.

\begin{table*}
  \centering\smaller
  \tabcolsep 0.35em
  \begin{tabular}{lrrrrrrrrrrrrrrrr}
    \toprule
    & \multicolumn{4}{c}{\textbf{\HPLT}}
    & \multicolumn{4}{c}{\textbf{FineWeb 1.4.0, 2.1.0}}
    & \multicolumn{4}{c}{\textbf{HPLT 2.0}}
    & \multicolumn{4}{c}{\textbf{MADLAD-400 1.0}}\\
    \textbf{Language}
    & \multicolumn{1}{c}{\textbf{D}} & \multicolumn{1}{c}{\textbf{T}}
    & \multicolumn{1}{c}{\textbf{|\hskip 4pt|}} & \multicolumn{1}{c}{\textbf{\%}}
    & \multicolumn{1}{c}{\textbf{D}} & \multicolumn{1}{c}{\textbf{T}}
    & \multicolumn{1}{c}{\textbf{|\hskip 4pt|}} & \multicolumn{1}{c}{\textbf{\%}}
    & \multicolumn{1}{c}{\textbf{D}} & \multicolumn{1}{c}{\textbf{T}}
    & \multicolumn{1}{c}{\textbf{|\hskip 4pt|}} & \multicolumn{1}{c}{\textbf{\%}}
    & \multicolumn{1}{c}{\textbf{D}} & \multicolumn{1}{c}{\textbf{T}}
    & \multicolumn{1}{c}{\textbf{|\hskip 4pt|}} & \multicolumn{1}{c}{\textbf{\%}}\\
    \cmidrule(r){1-1} \cmidrule(rl){2-5} \cmidrule(rl){6-9} \cmidrule(rl){10-13} \cmidrule(l){14-17}
    \textbf{English}
    & 18B & 16T & 901 & 55
    & 24B & 17T & 695 & 78
    & 4.4B & 3.9T & 892 & 35
    & 1.5B & 1.7T & 1099 & 38\\
    \textbf{Multilingual}
    & 11B & 13T & 1187 & 45
    & 5.0B & 4.9T & 976 & 22
    & 6.1B & 7.2T & 1178 & 65
    & 2.1B & 2.7T & 1266 & 62\\
    \cmidrule(r){1-1} \cmidrule(rl){2-5} \cmidrule(rl){6-9} \cmidrule(rl){10-13} \cmidrule(l){14-17}
    \textbf{Basque}
    & 3.2M & 3.2B & 991 & 0.02
    & 1.6M & 1.5B & 951 & 0.03
    & 2.0M & 2.0B & 1030 & 0.03
    & 1.6M & 1.5B & 1258 & 0.05\\
    \textbf{Catalan}
    & 2.6M & 22B & 853 & 0.17
    & 1.7M & 12B & 715 & 0.25
    & 1.9M & 18B & 976 & 0.25
    & 0.9M & 10B & 1084 & 0.38\\
    \textbf{Czech}
    & 107M & 126B & 1171 & 0.93
    &  66M &  67B & 1015 & 1.37
    &  75M &  95B & 1266 & 1.32
    &  38M &  50B & 1329 & 1.88\\
    \textbf{Finnish}
    & 49M & 73B & 1491 & 0.55
    & 36M & 48B & 1324 & 0.99
    & 34M & 53B & 1538 & 0.74
    & 20M & 34B & 1673 & 1.27\\
    \textbf{French}
    & 603M & 584B & 968 & 4.32
    & 360M & 292B & 811 & 5.95
    & 401M & 379B & 943 & 5.24
    & 216M & 269B & 12403 & 9.95\\
    \textbf{Galician}
    & 4.0M & 3.1B & 772 & 0.02
    & 2.5M & 1.8B & 695 & 0.04
    & 3.0M & 2.7B & 906 & 0.04
    & 0.2M & 1.3B & 1004 & 0.05\\
    \textbf{Norwegian}
    & 37M & 52B & 1388 & 0.39
    & 40M & 53B & 1318 & 1.09
    & 28M & 42B & 1477 & 0.58
    & 14M & 22B & 1508 & 0.83\\
    \textbf{Spanish}
    & 725M & 658B & 908 & 4.86
    & 441M & 329B & 746 & 6.71
    & 503M & 471B & 936 & 6.51
    & 250M & 254B & 1015 & 9.43\\
    \textbf{Ukrainian}
    & 80M & 81B & 1014 & 0.60
    & 53M & 49B & 938 & 1.02
    & 47M & 60B & 1280 & 0.84
    & 24M & 31B & 1268 & 1.17\\
    \bottomrule
  \end{tabular}
  \caption{Key statistics for \HPLT, contrasted with FineWeb, HPLT~2.0,
    and MADLAD-400.
    Columns show \textbf{D}-ocument and \textbf{T}-oken
    counts, as well as average document length (``\textbf{|\hskip 4pt|}'')
    and the token share of each subset of the total (``\textbf{\%}''); individual
    per-language percentages are of the non-English parts only.%
    \protect\footnotemark}
  \label{tb:overall-statistics}
\end{table*}

\paragraph{Annotation} 
%
%
The Web Docs Scorer (WDS) approach\footnote{\url{https://github.com/pablop16n/web-docs-scorer}}
provides an integrated consolidation of a rich tradition of heuristic document
filtering, e.g.\ based on length statistics, presence of ``oddity'' signals,
and document- vs.\ segment-level language distributions.
We adapt and refine this tool and make WDS levels a central annotation in
the \HPLT\ dataset (see below).
Furthermore, we update and re-train the Turku web register classifier
\cite{Henriksson.etal.2024} and apply register annotations for 104 of
the languages in \HPLT.

\paragraph{Packaging} 

The WDS scores described above provide a concise and 
language-agnostic perspective on document properties that we expect to
correlate with their prospective utility in LLM development.
For flexible experimentation with WDS-based data sampling from \HPLT, we bin
documents within each language by WDS levels --
$\{$5,\,6,\,7,\,8,\,9,\,10$\}$ -- and globally sort each bin.
\sref{ss:hplt-e-evaluation} below presents experimental evidence that training
on higher WDS levels can enhance LLM performance.
All \HPLT\ data -- documents and metadata -- is packaged in
Zstandard-compressed JSONlines form, where larger WDS bins are broken up into 
multiple shards. 
The total release comprises some 50 terabytes in about 3000 files, which are
distributed via HTTP download.

\section{Overall Dataset Statistics} 
\label{sc:statistics}

To put the \HPLT\ monolingual dataset into perspective,
Table~\ref{tb:overall-statistics} presents document and token
counts\footnote{For the purpose of comparable statistics across languages and
different datasets, token counts are computed using the Gemma-3 tokenizer
\cite{Gemma-3}, a SentencePiece tokenizer with a vocabulary of 256K
sub-words and good coverage for all target languages.}
for the English and multilingual (`non-English') partitions of the data, as
well as counts for a small sample of individual languages.
For ease of comparison, these statistics are accompanied with average document
lengths and per-language proportions, and contrasted with corresponding figures 
for three other publicly available multilingual datasets mentioned in
\sref{sc:introduction}.
\footnotetext{All dataset references in the remainder of this paper are to the
  specific versions indicated in the table headers.}

As is evident from these numbers, \HPLT\ is the by far largest
publicly available such dataset, and its multilingual breadth compares
favorably to other widely used resources.
In Gemma-3 tokens, the multilingual \HPLT\ partition is about 2--3 times larger
than FineWeb and HPLT~2.0, respectively, and five times larger than the older
MADLAD-400 dataset.
In terms of average document length, \HPLT\ patterns with HPLT~2.0, markedly ahead
of FineWeb and well behind MADLAD-400.
A small selection of European languages in Table~\ref{tb:overall-statistics}
shows languages ranging between a ``mere'' billion of available tokens to
others with hundreds of billions.

\section{In-Depth Analytics} 
\label{sc:analytics}
%
%
Like \citet{burchell-etal-2025-expanded}, we calculate descriptive statistics
for \HPLT\ using the HPLT Analytics
tool\footnote{\url{https://github.com/hplt-project/data-analytics-tool}} and
compare \HPLT\ to their dataset.\footnote{Graphics and examples for this
analysis are provided in Appendices~\ref{ax:stats},
\ref{ax:domains}, \ref{ax:tlds}, and \ref{ax:ngrams}.}

\paragraph{Descriptive Statistics} We notice a substantial difference in unique
segments, 73\% in \HPLT\ vs.\ 52\% in HPLT~2.0, on average. This likely reflects
global rather than per-crawl deduplication (see above) that removes near-identical
documents, thus increasing text diversity.
The difference is in general higher for small-to-medium languages
and lower in larger datasets, which generally tend to have a smaller proportion of
unique segments.
However, the ratios of large documents (above 25 segments) and short segments
(less than 3 tokens) show no significant differences: 21\% vs. 20\% for large documents and 13.1\% vs. 11.5\% for short segments (on average), for \HPLT\ and 2.0, respectively.
Similarly, proportions of individual segments in the document language
are comparable, at 71.5\% vs.\ 71.9\%, on average.
Lower values in this metric are usually associated with low resourced languages
or languages easily confused with similar, more-resourced ones.

\paragraph{Internet Domains}
Compared to HPLT~2.0 and because of the global deduplication process,
\HPLT\ exhibits a wider variety of domains.
For example, we notice that while HPLT~2.0 has more than 20 datasets with at least
25\% of their documents coming from Wikipedia pages, this holds for only seven
datasets in \HPLT\ (Wikipedia documents in HPLT~2.0 are likely to be repeated
across collections, thus overrepresented in the dataset).
However, and similar to HPLT~2.0, biblical webpages are a large part of the content
of smaller language datasets (especially African languages), with up to 25
datasets in \HPLT\ with more than half of their documents originating from this
kind of domains. 

Regarding Top Level Domains (TLDs), we find that the geographic TLD
distribution in \HPLT\ is similar to HPLT~2.0, usually with the most frequent
geographic TLD corresponding to the country where the language of the dataset
is spoken.
This ratio tends to be higher in mid-sized European languages.
Similarly, we find languages containing geographic TLDs from various countries
where its language is spoken (suggesting the proportions of different variants
of the language in the dataset) and some datasets showing a variety of
geographic TLDs from closely related countries or territories, indicating
possible deficiencies with language identification. 

\paragraph{N-grams}
HPLT Analytics calculates the five most common $n$-grams for orders 1 to 5 in each dataset, after discarding $n$-grams that start or end with a stopword.
Regarding frequent $n$-grams, the most pronounced difference when comparing to
HPLT~2.0 is the lack of Wikipedia-related $n$-grams (see above).
We also find indicators of low-quality text in some of the datasets that are
more prominent in \HPLT\ than in 2.0, such as adult (notably in European
languages) or betting content. 

\paragraph{Register Labels}
After inspecting the distribution of web registers (see above) among the different languages of \HPLT\ for which they are available, we find that the most common register tends to be \textit{Narrative}, especially the \textit{News Report} subregister.
However, for larger European languages, the \textit{Narrative} register tends to be tied or surpassed by \textit{Information Persuasion}, noticeably the \textit{Description with Intent to Sell} subregister.
In HPLT~2.0, the same two registers appear the most frequent, but in this case the most frequent subregister labels are \textit{Narrative Blog} and \textit{Encyclopedia Article}.
This contrast in blogs and encyclopedic articles documents is likely also attributable to global deduplication.

\section{Manual Inspection} 
\label{sc:inspection}

For 23 languages, native or fluent speakers have manually inspected randomly sampled documents and marked those that contain pornographic content, text with artifacts (e.g.\ navigational elements, headings or list items without proper delimitation, truncated text, or snippet markers), unnatural text (e.g.\ word lists for search engine optimization, high proportions of ``boilerplate'', or very obvious machine translation), or incorrectly identified language. 
Reflecting availability of annotators, for each language between 50 and 1000 documents were inspected, on average 348 documents per language. 
Table~\ref{tab:manual_inspection} reports confidence intervals for percentages of problematic texts of different types.\footnote{The inversion of the binomial test is employed to calculate confidence intervals.}  

\begin{table}[]
  \smaller
  \renewcommand{\arraystretch}{0.95}
\tabcolsep 0.45em
\centering
\begin{tabular}{lcccc}
\toprule
\textbf{Language} & \textbf{Porn}  & \textbf{Artifacts} & \textbf{Unnatural} & \textbf{LID} \\
\midrule

Asturian                      & 0-1                               & 2-9                                         & 2-8                                    & 19-31                                       \\
Bosnian                       & 0-0                               & 52-62                                       & 3-8                                    & 66-75                                       \\
Catalan                       & 0-1                               & 0-3                                         & 0-1                                    & 0-2                                         \\
Chinese                       & 0-3                               & 0-5                                         & 2-8                                    & 0-2                                         \\
Croatian                      & 0-2                               & 42-52                                       & \phantom{0}7-14                                   & \phantom{0}6-12                                        \\
Czech                         & 0-1                               & 4-7                                         & 24-30                                  & 0-1                                         \\
English                       & 0-2                               & 29-43                                       & 0-5                                    & 0-1                                         \\
Finnish                       & 2-4                               & 15-20                                       & \phantom{0}7-11                                   & 0-0                                         \\
French                        & 0-4                               & 13-23                                       & 10-21                                  & 0-1                                         \\
Galician                      & 0-1                               & \phantom{0}4-11                                        & 0-5                                    & 0-3                                         \\
German                        & 0-1                               & 0-3                                         & \phantom{0}5-13                                   & 0-1                                         \\
Hindi                         & 0-1                               & 20-33                                       & 1-6                                    & 0-4                                         \\
Iranian Persian               & 0-1                               & 35-45                                       & \phantom{0}8-15                                   & 1-4                                         \\
Italian                       & 0-4                               & \phantom{0}4-11                                        & \phantom{0}5-14                                   & 0-2                                         \\
Japanese                      & 0-2                               & 47-61                                       & 13-23                                  & 0-1                                         \\
Modern Greek         & 0-1                               & 31-37                                       & 2-5                                    & 0-1                                         \\
Norwegian Bokmål              & 2-7                               & 12-20                                       & 2-7                                    & 0-3                                         \\
Portuguese                       & 0-2 	 & 30-44 	& \phantom{0}4-11 	& 0-3                                 \\
Russian                       & 0-1                               & 4-9                                         & 2-5                                    & 0-1                                         \\
Serbian                       & 0-6                               & 48-75                                       & \phantom{0}6-27                                   & 0-6                                         \\
Slovak                        & 0-0                               & 11-19                                       & 4-9                                    & 0-2                                         \\
Spanish                       & 0-1                               & 15-23                                       & \phantom{0}5-10                                   & 0-1                                         \\
Yoruba                        & 0-3                               & \phantom{0}7-16                                        & 0-5                                    & \phantom{0}4-11                                        \\
\end{tabular}
\caption{Manual inspection of the \HPLT{} dataset samples: 95\% confidence intervals for the percentages of texts containing porn, artifacts, unnatural texts, and texts in a different language (LID errors).}
\label{tab:manual_inspection}
\end{table}

We observe that the proportion of porn in the dataset is below 2\% for most languages. The proportion of the documents where the  language is mis-identified is also quite low, with the notable exceptions of the Bosnian dataset, which consists mostly of texts in Serbian, and the Asturian dataset, which often contains Spanish. The proportions of unnatural texts and texts containing artifacts widely vary across languages. This is related in part to the fuzzy definition of these properties in the guidelines and subjectivity in judgments. For example, among two annotators of Czech, one annotated 34\% of text as unnatural while another only 1\%.
The notion of text naturalness is subjective by definition.

\section{Multilingual LLM Evaluation} 
\label{sc:evaluation}

We develop HPLT-e\footnote{\url{https://github.com/hplt-project/hplt-e}}, a framework for automated large-scale multilingual evaluation designed to systematically compare and refine data preparation choices across nine selected languages shown in Table~\ref{tb:overall-statistics}. These languages are chosen to ensure both availability of native speakers in our development team and a minimum level of diversity in terms of language resources, families, and scripts. Similar to \citet{penedo2024fineweb,penedo2025fineweb2pipelinescale}, we pretrain separate language models per language using an otherwise fixed pretraining setup, and evaluate them at regular checkpoint intervals (every 1B tokens) in a 0-shot regime, carefully selecting tasks that meet the pretraining evaluation signal criteria below.

\subsection{HPLT-e: Framework}
\label{ss:hplt-e-evaluation}
HPLT-e includes 127 language understanding and generation tasks, each supporting 3--7 human-written prompts. We aim to cover different task categories in all languages:\ entailment, commonsense reasoning, language-specific \& world knowledge, paraphrasing, reading comprehension, sentiment analysis, toxicity detection, and truthfulness. HPLT-e integrates with LM Evaluation Harness \cite{eval-harness}, for experimental flexibility and replicability.
\paragraph{Pretraining}
We pretrained  decoder-only models of the size 2.15B on 100B tokens sampled from FineWeb, HPLT 2.0, MADLAD-400, and \HPLT.\footnote{For lower-resource languages with less than 100B tokens of available data, datasets are uniformly upsampled (repeated) following 
\citet{muennighoff2023scaling}.} All models employ the Gemma-3 tokenizer and follow the Llama architecture \cite{touvron2023llama} with 24 layers, 32 attention heads, and
a sequence length of 2048. Pretraining is run using the Megatron-LM \cite{megatron-lm} framework on the \href{https://www.lumi-supercomputer.eu}{LUMI}
supercomputer, utilizing 16 nodes with AMD MI250x GPUs for a total of
approximately 3,000 GPU hours per model.
The estimated carbon footprint per model is 59 kg CO\textsubscript{2}. 

\paragraph{Benchmarks} We combine open-source human-curated multi-task benchmarks: IberoBench (Catalan, Spanish, Basque, Galician; \citealp[]{baucells-etal-2025-iberobench}), FrenchBench (French; \citealp[]{faysse2024croissantllm}), NorEval (Norwegian Bokmål and Nynorsk; \citealp[]{mikhailov-etal-2025-noreval}), BenCzechMark (Czech; \citealp[]{fajcik2025benczechmark}), and \href{https://github.com/LumiOpen/lm-evaluation-harness/tree/finbench_v2/lm_eval/tasks/finbench_v2}{Finbench v2} built on Finbench (Finnish; \citealp[]{luukkonen-etal-2023-fingpt}). In addition, we create a benchmark for Ukrainian (UkrainianBench), which comprises Global MMLU \cite{singh-etal-2025-global}, INCLUDE \cite{romanou2025include}, UA-SQuAD \cite{ua_datasets_2021}, ZNO \cite{romanyshyn-etal-2024-unlp}, Belebele \cite{bandarkar-etal-2024-belebele}, TextDetox \cite{dementieva-etal-2024-toxicity}, and WMT24++ \cite{deutsch-etal-2025-wmt24}. 

\paragraph{Prompt Collection} HPLT-e enables evaluation across 500+ prompts that have diverse structure and answer formatting to mitigate prompt sensitivity, a model limitation where variations in prompt formulation can affect downstream performance \citep[e.g.][]{pezeshkpour-hruschka-2024-large,sclar2024quantifying}. We adapt single-prompt benchmarks (IberoBench, FrenchBench, and UkrainianBench) to the multi-prompt design through (i) manual translation of English prompts from PromptSource \cite{bach-etal-2022-promptsource} and (ii) prompt creation by native speakers.

\paragraph{Task Selection} We use standard task-specific metrics and report the maximum score across the prompts as the main aggregation method. We extend the FineTasks evaluation design \cite{penedo2025fineweb2pipelinescale} and select tasks that provide pretraining evaluation signal based on the following key criteria: \textit{monotonicity} -- the model performance should improve as pretraining progresses; \textit{stable pretraining} -- the relative variability of model performance across pretraining intervals should be low; \textit{non-randomness} -- the final model performance should exceed a random guessing baseline and be satisfactory; \textit{ranking consistency} -- the relative ranking of models across pretraining intervals should remain stable; \textit{low noise} -- the final model performance should be high relative to task variability across prompts; \textit{low prompt sensitivity} -- the median absolute deviation across prompts should be low to ensure prompt-invariant dataset comparison; \textit{prompt lottery} -- the frequency with which the best-performing prompt changes as pretraining progresses should be low. 

\begin{figure}[t!]
  \centering
  \includegraphics[width=\columnwidth]{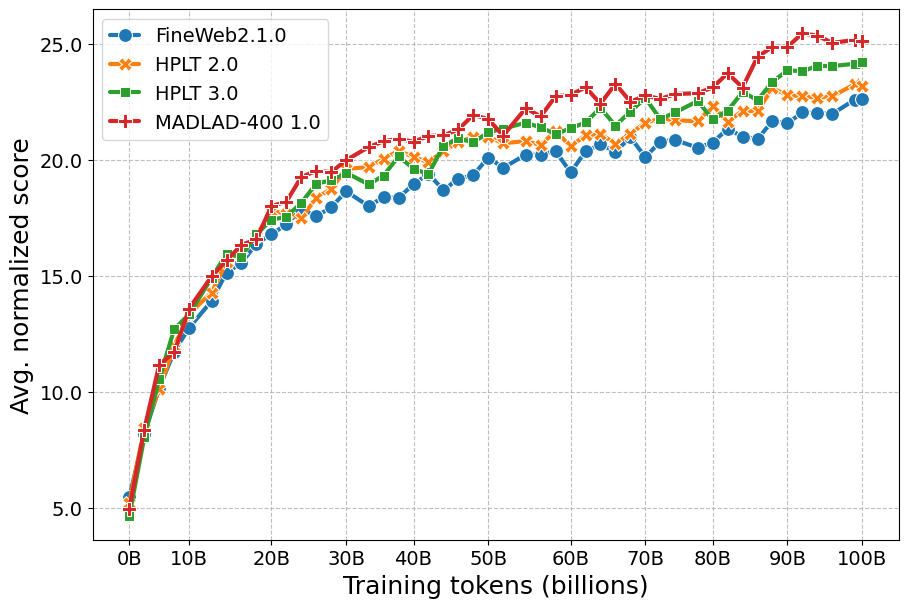} 
  \caption{Comparison of models pretrained on FineWeb, HPLT 2.0, 3.0, and MADLAD-400.}
  \label{fg:corpora_comparison}
\end{figure}

\paragraph{Performance Aggregation} Following \citet{fourrier2024open,penedo2025fineweb2pipelinescale}, we compute a \textit{language score} as the average of min--max-normalized performance scores across selected tasks. In particular, we first rescale all scores between the random baseline and the maximum possible score. We then average the rescaled scores within each task category and take the average of per-category scores to compute the language score. To produce a \textit{multilingual score}, we utilize several approaches: \textit{average language score} -- we compute the average of all language scores; \textit{average multilingual rank} -- we rank the final models' language scores across all datasets and average their ranks; \textit{Borda's count} -- using Vote'n'Rank \cite{rofin-etal-2023-votenrank}, we compute the final models' Borda rankings within each language and aggregate the per-language rankings to produce the overall ranking. Borda's count serves as an alternative to average-based aggregation, allowing for aggregating heterogeneous metrics by leveraging rank-based differences among models.


\subsection{Results} 
In line with \citet{penedo2025fineweb2pipelinescale}, we find that tasks for lesser-resourced languages, notably Basque and Galician, are unsuitable for pretraining evaluation due to potential difficulty and lack of monotonic performance progression during pretraining. We thus report our key findings on a final suite of 26 selected tasks across seven remaining languages.

\paragraph{Dataset Comparison} Figure~\ref{fg:corpora_comparison} presents the results of comparing the models pretrained on different datasets. All models show a monotonic performance improvement on our selected tasks as pretraining progresses. Models pretrained on MADLAD-400 achieve the highest multilingual score, followed by \HPLT, while HPLT~2.0 and FineWeb perform on par. 

These results are consistent with rank-based aggregation. Models are ranked as (1) MADLAD-400; (2) \HPLT; and (3) HPLT~2.0 and FineWeb; by average multilingual ranks, HPLT~2.0 slightly outperforms FineWeb, whereas Borda's counts show the inverse ordering. Overall, our findings indicate that refined data preparation in \HPLT\ has improved average dataset quality, which translates into competitive performance gains for model pretraining.
The performance of models trained on the older and smaller MADLAD-400 data calls for follow-up studies.

\begin{figure}[t!]
  \centering
  \includegraphics[width=\columnwidth]{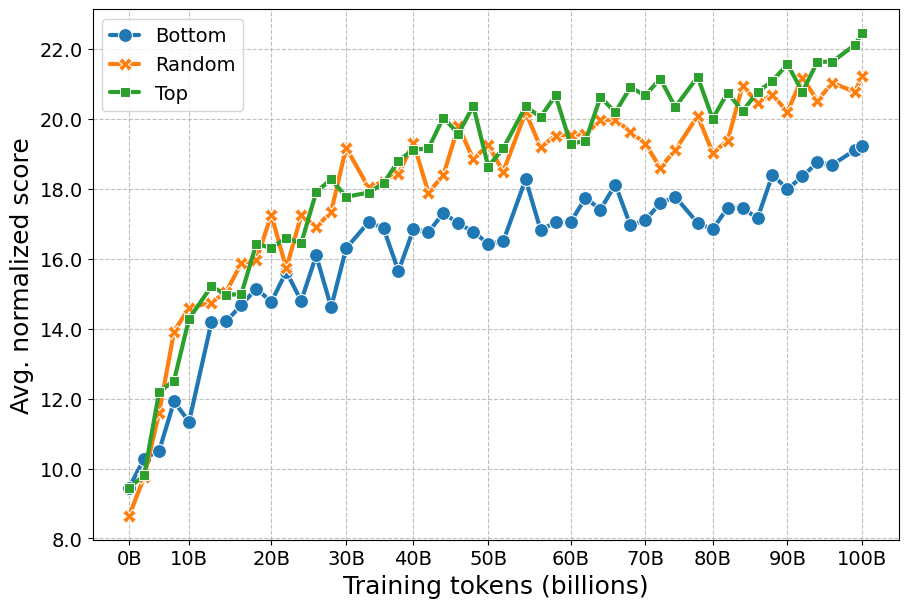} 
  \caption{Comparison of different WDS-based sampling strategies from the Spanish \HPLT\ data.}
  \label{fg:spa_Latn_wds_comparison}
\end{figure}

\paragraph{Sampling by Quality Estimates} 
As another application of HPLT-e, Figure~\ref{fg:spa_Latn_wds_comparison} shows
the performance of three Spanish models trained on different subsets of
\HPLT\ training data selected according to WDS levels; see \sref{sc:processing}
above.
In these experiments, five Spanish tasks were selected as informative for all
models.
Here, ``random'' sampling represents the default approach, drawing uniformly on
the full dataset, while ``top'' and ``bottom'' take advantage of the sorting by
WDS levels and sequentially draw 100B training tokens from either end of the
dataset.
Low WDS levels clearly lead to inferior model performance, while sampling from
only the ``top'' does not clearly improve over the full dataset, possibly owing
to overly limited diversity.
We observe similar patterns in French, the other language supported by HPLT-e
where there are hundreds of billions of available training data.

\section{Monolingual Encoders} 
\label{sc:gpt-bert}
We trained monolingual encoder GPT-BERT language models
\cite{charpentier-samuel-2024-bert} on near 60 \HPLT\ languages.
The GPT-BERT architecture combines masked and causal language modeling within a
single Transformer stack. 

\paragraph{Selection of Languages} 
\label{ss:t-five-languages}
Our choice of languages to train encoder models (as well as encoder--decoder
models; see \sref{sc:t-five} below) on was motivated by their typological
diversity.
This means we aimed at covering as many different language families as
possible. At the same time, we did not train models on extremely
under-represented languages (less than $\approx$0.25M documents in our
datasets).
As a result, our models cover the following language families:\ Indo-European,
Sino-Tibetan, Japanese, Austronesian, Austro-Asiatic, Uralic, Altaic,
Afro-Asiatic, Korean, Tai-Kadai, Dravidian, Kartvelian, Niger-Kongo, and
Basque.

\paragraph{Evaluation}
We evaluated the GPT-BERT models on the standard Universal Dependencies tasks
(PoS-tagging, lemmatization, dependency parsing) and on named entity
recognition (NER) using the WikiAnn benchmark
\cite{rahimi-etal-2019-massively}.
Appendix~\ref{ax:models} provides detailed performance comparisons for
our GPT-BERT models to mBERT \cite{devlin-etal-2019-bert}, XLM-R
\cite{conneau-etal-2020-unsupervised}, mmBERT
\cite{marone2025mmbertmodernmultilingualencoder}, and ``classic'' BERT models trained on the
previous HPLT data releases (see Table~\ref{tab:mlm}).
For the majority of languages and tasks, HPLT 2.0 or \HPLT\ models are the
optimal choice, but we recommend to check the performance for specific
language--task combinations. 

In addition, we evaluated linguistic competence of our GPT-BERT models using
MultiBLIMP \cite{multiblimp}.
This is a dataset of minimal pairs: Each sample consists of a pair of
sentences.
The first sentence is grammatical (`correct'). The second sentence is identical to the first one, but its syntactic head is modified to make the resulting sentence ungrammatical (`wrong'). We re-use MultiBLIMP's official evaluation code, adding support for masked language models \cite{misra2022minicons, kauf-ivanova-2023-better}. Regardless of the particular model architecture, evaluation works as follows: Probabilities of both `correct' and `wrong' sentences are produced using the language model under evaluation. The resulting accuracy is the share of samples where the probability of the `correct' sentence is higher than that of the `wrong' one. 
Since Norwegian is missing from MultiBLIMP, we used the grammatical error correction benchmark NoCola \cite{jentoft-samuel-2023-nocola}. Unlike MultiBLIMP, its samples may contain more than one ungrammatical subsequence (derived from real Norwegian as a second language learners' mistakes), and it may be not only words, but also punctuation marks, e.g.\ a single comma. 

Interestingly, on MultiBLIMP, \HPLT\ GPT-BERT models outperform  pretty much
all the baselines: XLM-R, mmBERT and BERT models trained on previous HPLT
releases (see Table~\ref{tab:mlm-multiblimp} in Appendix~\ref{ax:models}).
When evaluated as causal language models, GPT-BERTs also consistently outperform Goldfish, which was the best system for many languages in the original MultiBLIMP paper \cite{multiblimp}.

\HPLT\ GPT-BERT models are publicly available, including intermediate checkpoints.\footnote{\url{https://hf.co/collections/HPLT/hplt-30-gpt-bert-models}}

\section{Monolingual Encoder--Decoders} 
\label{sc:t-five}

In addition to encoder-only models, we used the \HPLT\ dataset to train and evaluate 57 language-specific monolingual \textit{encoder--decoder} language models, following the T5-base architecture \cite{JMLR:v21:20-074, samuel-etal-2023-norbench}.
This novel family of models, including intermediate checkpoints, is also publicly available.\footnote{\url{https://hf.co/collections/HPLT/hplt-30-t5-models}}

\paragraph{Motivation \& Approach} 
\label{ss:t-five-setup}

 Despite the popularity of decoder-only LLMs in recent years,
 encoder--decoders are still widely used in real-world applications, showing
 strength in both generative and discriminative tasks \cite{zhang2025encoderdecodergemmaimprovingqualityefficiency}.

Our T5-like models serve two primary purposes:
\begin{enumerate}
    \item to evaluate \HPLT{} quality as training data
      across a large number of languages; and
    \item to provide a family of comparable monolingual encoder--decoders
      trained on current data.
\end{enumerate}

\begin{table*}[t!]
  \centering\smaller
  \tabcolsep 0.3em
  \begin{tabular*}{\textwidth}{@{\extracolsep{\fill}}lrccrccc}
    \toprule
    \textbf{Language} 
    & \multicolumn{3}{c}{\bf Named Entity Recognition \textbf{(WikiAnn, F1)}}
    & \multicolumn{4}{c}{\bf Linguistic Competence (\textbf{MultiBLIMP, Acc)}}\\
     & \textbf{size} & \textbf{T5 \HPLT} &
     \textbf{mT5-base} &
    \textbf{size} & \textbf{T5 \HPLT} 
    & \textbf{mT5-base} & \textbf{mT5-xxl} \\
    \cmidrule(r){1-1}\cmidrule(rl){2-4}\cmidrule(l){5-8}
    Catalan (\textbf{cat\_Latn}) & 10000 & 92.7 & 87.4 & 2284 & 95.6 & 91.6 & 93.0 \\
    Czech (\textbf{ces\_Latn}) & 10000 & 91.6 & 85.2 & 4256 & 95.9 & 88.8 & 93.4 \\
    English (\textbf{eng\_Latn}) & 10000 & 82.1 & 77.6 & 770 & 94.2 & 90.6 & 95.3 \\
    Basque (\textbf{eus\_Latn}) & 10000 & 92.0 & 82.8 & 273 & 97.4 & 94.9 & 96.0 \\
    Finnish (\textbf{fin\_Latn}) & 10000 & 90.3 & \phantom{0}1.8 & 2570 & 95.6 & 81.4 & 86.1 \\
    French (\textbf{fra\_Latn}) & 10000 & 88.9 & 83.3 & 2548 & 93.6 & 91.7 & 94.8 \\
    Galician (\textbf{glg\_Latn}) & 10000 & 93.4 & 89.2 & 753 & 96.0 & 90.7 & 95.4 \\
    Bokmål (\textbf{nob\_Latn}) & 10000 & 93.2 & 87.0 & \textbf{*}3463 & 40.6 & 68.0 & 71.8 \\
    Nynorsk (\textbf{nno\_Latn})  & 1000 & 94.0 & 88.2 & - & - & - & - \\
    Spanish (\textbf{spa\_Latn})  & 10000 & 90.7 & 84.0 & 2541 & 95.2 & 93.8 & 96.3 \\
    Ukrainian (\textbf{ukr\_Cyrl}) & 10000 & 92.5 & 82.1 & 2744 & 95.7 & 89.4 & 94.8 \\
    \cmidrule(r){1-1}\cmidrule(rl){2-4}\cmidrule(l){5-8}
    \textbf{Average} & - & \textbf{90.5} & 78.8 & - & \textbf{93.5} & 86.8 & 91.4 \\
    \bottomrule
  \end{tabular*}
  \caption{Evaluation results of \HPLT{} monolingual encoder--decoders (11 languages of interest, Bokmål and Nynorsk are two varieties of Norwegian), along with the test set sizes for each language. Average results are computed over all the 57 languages we have trained models for (see full results in Appendix~\ref{ax:models}). \textbf{*}Bokmål competence benchmarks are not part of MultiBLIMP.}
  \label{tb:t5}
\end{table*}

As regards the second purpose, we note the only available encoder--decoder
with multilingual capabilities is mT5 pretrained on the mC4 corpus
\cite{xue-etal-2021-mt5}, and its instruction-tuned derivatives,
e.g.\ mT0 \cite{muennighoff-etal-2023-crosslingual} and Aya \cite{ustun-etal-2024-aya}.
Our collection of T5-like models is trained on more recent \HPLT{} datasets,
which we believe to be substantially cleaner and more high-quality than mC4
(see our evaluation below).
Thus, we hope that the NLP community will benefit from these novel models.
In addition, language-specific models can be used to conduct comparative
studies of LLM learning process.
Their size ($\approx$275M parameters each) allows for easy and
computationally cheap deployment.


\paragraph{Evaluation} 
\label{ss:t-five-evaluation}

We evaluate our T5-like models on two tasks: (i) named entity recognition with WikiAnn, (ii) linguistic competence with MultiBLiMP (NoCOLA for Norwegian). Their performance was compared to the performance of the original mT5-base\footnote{\url{https://hf.co/google/mt5-base}}
 model. We also demonstrate the performance of a much larger mT5-xxl\footnote{\url{https://hf.co/google/mt5-xxl}}
 model on the MultiBLiMP benchmark. 

We re-use MultiBLIMP's official evaluation code, adding support for encoder--decoder models. 
The difference between  decoder-only and encoder--decoder models is that with the former, each token is only conditioned on the previous tokens and a sentence may be evaluated `as is', while with the latter, each token is conditioned both on the encoder's output and the previous tokens; thus, we replicate the T5 training procedure and mask the syntactic head in the encoder input, and the rest of tokens in the decoder input. 


Table~\ref{tb:t5} shows the selected evaluation results for the languages of interest as defined in \sref{sc:evaluation} plus English. It demonstrates that the \HPLT{} monolingual encoder--decoders offer a competitive alternative to the multilingual mT5 models, outperforming mT5-base on the NER task, and outperforming \textit{both} mT5-base and mT5-xxl on MultiBLIMP.

We also conducted additional evaluations on the English MultiBLIMP: The original monolingual T5-base performed better (93.5\% accuracy) than the multilingual one; instruction-tuned derivatives \texttt{mt0-xl} 
 (90.5\% accuracy) and \texttt{aya-101} 
 (86.6\% accuracy) further proved the finding by \citet{multiblimp} that fine-tuning tends to worsen BLIMP performance; the most recent English T5-base model, \texttt{t5gemma-b-b-ul2}, 
showed surprisingly low result (66.5\% accuracy); however, its training objective was more sophisticated \cite{tay2023ul} than in the original T5. We leave further adjusting our inputs to it for future work.

\section{Mining for Bilingual Texts} 
\label{sc:bitexting}
After constructing and evaluating the monolingual corpus, a natural next step
is to further leverage these resources to mine parallel data.
Although the field is increasingly favoring LLMs over traditional machine
translation (MT) encoder–decoder architectures, the use of parallel corpora in
LLM pretraining has been shown to enhance multilingual capabilities
\cite{zhang2024enhancing,qorib-etal-2025-just}.
Moreover, MT systems can be used to generate synthetic data from monolingual
resources that can be useful for LLM pretraining, as discussed in
\sref{sc:multisynt}.

We release parallel data at both sentence- and document-level. Recent work in MT is moving towards document-level translation. This shift is reflected in the release of training datasets \cite{merx2025openwho,obrien2025dochplt}, dedicated benchmarks \cite{fernandes-etal-2023-translation,deutsch2025wmt24++}, and the development of targeted training strategies \cite{ramos2025multilingual}.
This is motivated by the ability of LLMs to leverage long context, which allows them to model discourse phenomena such as coreference resolution, anaphora or ellipsis, which are typically lost when restricting training to the sentence level.

We create English-centric parallel resources for 28 European languages (see Table~\ref{tab:bitext-stats} in the appendix for details), diverse in terms of linguistic families and resource availability.
We use Bitextor\footnote{\url{https://github.com/bitextor/bitextor}} for parallel data extraction. Bitextor takes as input the cleaned monolingual documents. The pipeline performs sentence segmentation, translates sentence into synthetic English for document alignment, and then applies cleaning rules and deduplication. We follow \citet{de-gibert-etal-2024-new,burchell-etal-2025-expanded} with minor modifications.
The main change we apply is that we maintain document structure from the beginning of the extraction.

Altogether, we extract roughly 1.1 billion sentence pairs aligned (after the final filtering step using bicleaner-AI), ranging from about 350K sentence pairs for the smallest bitext (Norwegian Nynorsk--English) to over 120 million sentence pairs (for Danish--English). To verify the usefulness of the data, we also trained baseline NMT models using Marian-NMT and Transformer-base models. Automatic evaluation scores using BLEU, ChrF and COMET over the FLORES200 devtest benchmark are also provided in appendix in Table~\ref{tab:mt-models}. Note that we did not perform any particular optimization of the training procedures not did we apply any other common techniques like data augmentation with back-translation etc. Nevertheless, the scores reveal that the extracted data provide a valuable resource out-of-the-box enriching the open resources of parallel data for the selected language pairs.

Furthermore, we also release the document-level parallel data including a multilingually aligned corpus across all languages using the English documents as a pivot. The strategy to keep document information enabled a straightforward alignment across language pairs producing parallel document-level corpus spanning 406 language pairs including roughly 230 million documents and 182 billion words altogether. The corpus is available from OPUS.\footnote{\url{https://opus.nlpl.eu/datasets/DocHPLT}}



\section{MT for Synthetic Data Generation} 
\label{sc:multisynt}

Many studies have shown the value of synthetic data, for example, \cite{doshi-etal-2024-pretraining,wang2025multilinguallanguagemodelpretraining}
and in this work we explore the use of machine translation as an effective short-cut to transfer knowledge from a resource-rich source language to under-resourced target languages~\cite{degibert2025scaling}. 

Many open-source models are available and can easily be integrated in translation workflows~\cite{tiedemann2023democratizing,nllbteam2022languageleftbehindscaling}.
For scalability, inference-efficient models are needed and, for that reason, we rely on compact encoder--decoder models instead of instruction-tuned LLMs. In particular, we select high-performing models available through the OPUS-MT Dashboard~\cite{tiedemann-de-gibert-2023-opus} and 
sample around 28 billion tokens of English data from FineWeb-Edu~\cite{lozhkov2024fineweb-edu} and 100 billion tokens from Nemotron-CC~\cite{su-etal-2025-nemotron} (taking the high-quality subset) to be translated. 
We split documents into sentences using the LoomchildSegmenter\footnote{\url{https://pypi.org/project/loomchild-segment/}}. Short segments are merged into multi-sentence segments until they exceed a maximum threshold of 1,024 characters to improve context-sensitivity.
We translate in shards of 500M tokens using Marian-NMT~\cite{mariannmt} and beam size four.
%
After translation, we merge translated sentences back into the original document structures,
leading to a document-level and segment-level aligned parallel dataset.
The data is published as part of the synOPUS collection\footnote{\url{https://opus.nlpl.eu/synthetic }} of synthetic parallel corpora. Furthermore, we also make the data available from Hugging Face\footnote{\url{https://hf.co/datasets/Helsinki-NLP/fineweb-edu-translated} and \url{https://hf.co/datasets/Helsinki-NLP/nemotron-cc-translated}} for seamless integration in common workflows.

FineWeb-Edu translations are available in 36 languages and language variants,\footnote{Language coverage (ISO-639-3 codes):\ bos, bul, cat, ces, dan, deu, ell, est, eus, fin, fra, gle, glg, hrv, hun, isl, ita, kat, lav, lit, mkd, mlt, nld, nno, nob, pol, por, ron, slk, slv, spa, sqi, srp, swe, tur, ukr.}
creating a corpus of around 1 trillion tokens. Translations from Nemotron-CC cover the same languages and 2.3 trillion tokens in total.

Initial experiments with small-scale language and translation models trained on
the synthetic data showed encouraging results on standard benchmarks indicating
potential value of automatic translation for data augmentation.
In our view, however, the prospective use of machine translations (or other
synthetic data) in LLM development calls for more in-depth studies -- for
example along the lines of \sref{sc:evaluation}, while emphasizing language
quality benchmarks and preferably looking at larger scales -- to tease
apart the candidate contributions, pitfalls, and ethical and legal risks of
training on ``non-natural'' data.

\section{Conclusions \& Outlook} 
\label{sc:conclusions}

\HPLT\ substantially refines an existing pipeline for very large-scale
preparation of mono- and bi-lingual datasets and applies it to a massive
collection of web archives.
All data, models, and software involved in this initiative are publicly
available under permissive terms of use.
The \HPLT\ dataset constitutes the largest multilingual resource of its type.
Contrastive experimentation suggests that its overall data quality enables LLM
performance superior to comparable resources.
This resource is supported by a novel multilingual evaluation framework and
accompanied by multiple families of pre-trained language models for a broad
range of languages, as well as by derived bilingual and synthetic text data.
Through this work, we hope to contribute to increased community activity
emphasizing generally available and transparent resources and processes for
very large-scale LLM research, with special attention to multilinguality and
linguistically and culturally adapted evaluation.

We invite community contributions to futher development, utlization, and
refinement of these resources.
In ongoing work, HPLT is preparing a fourth and final monolingual dataset
release, aiming to facilitate follow-up research and refinement.
The forthcoming HPLT 4.0 release will make available the complete document pool 
prior to deduplication, annotation, and filtering (see Figure~\ref{fg:pipeline}
above), close to 500 terabytes of compressed data.
Furthermore, annotations will be extended with predictions by multiple
data-driven ``quality'' classifiers, so as to enable fine-grained contrastive
experimentation with different quality signals and data sampling strategies.

\newpage
\section{Ethics Statement}
\label{sc:ethics}

Large-scale data curation from web archives is legally and ethically
challenging in a myriad of ways.
A large part of the content in these archives is expressly copyrighted or
published with license terms not intended for redistribution or use as training
material for e.g.\ language model or machine translation model development.
Another large portion of web content was originally published without explicit
terms of use or a specific license.

At the same time, massive web archives have long served as very valuable
sources of large-scale natural language data for a broad range of language
research and engineering tasks, where data volumes of at least (tens of)
billions of tokens often are prerequisites to much contemporary work.
Especially for medium- to low-resource languages (except for French and
Spanish, the majority of examples in Table~\ref{tb:overall-statistics} above),
there exist no alternative data sources on this scale.
These languages are under pressure in the digital age, and advancing language
technologies beyond the major language communities and markets can help
counterbalance these challenges.

The dilemma of responsible large-scale data curation from web archives is not
new, but it is, of course, greatly aggravated in the age of large language
models.
As sketched in \sref{sc:introduction} above, ``flagship'' data curation
initiatives like C4, FineWeb, MADLAD, or Nemotron-CC originate in corporate
environments from outside Europe.
The HPLT initiative presents a publicly funded EU effort to ``democratize'' the
web-scale data curation and LLM development landscape.
HPLT capitalizes on transparency and reuse of data, models, tools, and
knowledge to the highest possible degree.
In other words, our intentions are good and follow an ideal of generally
available web-scale resources as digital public goods.

The HPLT datasets are distillations from massive web crawls that are either
publicly available in and of themselves (in the case of the Common Crawl and
ArchiveBot), or have been provided to the consortium with the express consent
to redistribution of derived language resources by the original crawler (in the
case of the Internet Archive).
Nevertheless, neither the foundations who originally crawled the web, nor
anyone creating a derivative can assert ownership of these data.
Therefore, HPLT datasets are distributed with a dual licensing scheme,
where no rights are claimed or granted for the textual content per se (the
extracted text), while all metadata and annotations are put into the public
domain, using the Creative Commons CC0
license.\footnote{\url{https://creativecommons.org/public-domain/cc0/}}
From this point of view, it is the responsibility of the user of HPLT resources to
make sure their use is compliant with applicable legislation in their jurisdiction.
The HPLT download site makes this requirement clear as part of the terms of use
for the datasets.
This is combined with a low-threshold channel to submit take-down requests to
the consortium.

Additionally, document selection and filtering in HPLT put ample emphasis on
best practices to avoid redistribution of expressly prohibited or personally
sensitive content.
First, document selection from the Internet Archive collections retroactively
applies the de-facto standard \verb|robots.txt| opt-out mechanism, where web
sites can declare patterns that prohibit crawling.
As a conservative (or ``broad'') interpretation, the union of all
\verb|robots.txt| captures for a given site over the full duration of each
crawl -- typically multiple years, for the Internet Archive ``wide'' crawls --
are applied.
Second, in a subsequent filtering stage, all documents that are flagged by the
TruffleHog scanner for leaked
credentials\footnote{\url{https://github.com/trufflesecurity/trufflehog}} are
excluded from further processing.
Finally, for a broader protection of personal information, annotations on the
data include all matches from a state-of-the-art multilingual tool to detect
personally identifiable information
(PII)\footnote{\url{https://github.com/mmanteli/multilingual-PII-tool}}, for
example, email addresses, phone numbers, or IP addresses.
When used in LLM training, for example, it is expected that PII matches would
either be anonymized (masked out) or entire documents would be excluded from training
based on these annotations.

Speaking more broadly, there are of course numerous other ethical challenges in this line of work. 
Datasets curated from the web can contain all sorts of offensive or harmful
content, despite comprehensive efforts in the HPLT annotation and filtering
pipeline to clean the data.
Furthermore, beyond the general -- and substantial -- risks of bias propagation
and amplification, datasets for lower-resourced languages can be biased in
diverse ways.
Religious content, for example, often appears over-represented in our African
languages dataset.
While bias mitigation ultimately needs to be considered for each specific use
case and individual target language(s), we provide the HPLT Analytics framework
and detailed per-language analyses to help shed light on potentially
problematic properties in the HPLT datasets.

\newpage
\section{Limitations of this Work}
\label{sc:limitations}

In the limit, assessing the ``quality'' of trillions of tokens in text for
hundreds of languages is an impossible proposition.
The HPLT consortium has placed strong emphasis on data inspection from a
multitude of perspective, including those summarized in
\sref{sc:statistics}--\sref{sc:evaluation} above.
For multiple full release cycles, the project has applied quantitative
analysis, human data inspection, in-depth analytics, and ``extrinsic''
evaluation in terms of observed model performance when training on HPLT data.
However, all of these approaches, in turn, are subject to their own
methodological limitations, for example inevitable variation in human judgment
or limitations in automated downstream evaluation (for diverse languages).

Manifest limitations in, for example, the quality of text extraction, language
identification, removal of adult content, normalization and cleaning,
filtering on document quality signals, and others have been addressed through
incremental refinements of the HPLT pipeline.
Nevertheless, there are, of course, remaining limitations -- room for
improvement -- in all of these steps.
Seemingly mundane tasks like ``boilerplate'' removal (i.e.\ extracting the
``main'' content of a web document, while filtering out navigational elements
or advertisement) or document- and segment-level language identification are
challenging to optimize when dealing with very high degrees of variability and
``noise'' in the raw web data.
These challenges are often exacerbated for lower-resource languages or -- in
the case of language identification -- within linguistically closely related
families.

The HPLT 3.0 release is accompanied by interactive per-language in-depth
analytics reports and samples, which are intended to enable prospective users
of the data to perform focussed and comprehensive data analysis, to ultimately
gauge the utility of the data, including candidate shortages and risks, to a
specific use case.

\newpage
\section{Acknowledgments}
We thank Étienne Simon (UiO) and Daryna Dementieva (TUM) for their contribution to our prompt collection for French and Ukrainian and Erik Henriksson, Erofili Psaltaki, and Otto Tarkka (all UTU) for their contributions to the manual inspection. Furthermore, we would like to acknowledge the work done by language technology students at the University of Helsinki (Qing Li, Nirav Bhatt, Ilja Adel, Oona Itkonen, Tiankai Zang, Nikolay Vorontsov and Sopiko Kurdadze), running the bitext extraction pipeline and training reference translation models to test the extracted data.

This project has received funding from the Horizon Europe research and innovation programme of the European Union under Grant No.~101070350 and from UK Research and Innovation (UKRI) under the UK Horizon Europe funding guarantee, grant number 10052546. Final editing of the manuscript and its presentation have been supported by Grant No.~101195233 (Digital Europe programme of the European Union).
The contents of this publication are the sole responsibility of its authors and do not necessarily reflect the opinion of the European Union.
The authors wish to thank CESNET (Czech Republic), CSC (Finland) and Sigma2 (Norway) for computational resources and support.

\newpage
\section{Bibliographical References}
\label{sc:references}

\bibliographystyle{lrec2026-natbib}
\bibliography{anthology1,anthology2,references}

\begin{thebibliography}{64}
\expandafter\ifx\csname natexlab\endcsname\relax\def\natexlab#1{#1}\fi

\bibitem[{Bach et~al.(2022)Bach, Sanh, Yong, Webson, Raffel, Nayak, Sharma,
  Kim, Bari, Fevry, Alyafeai, Dey, Santilli, Sun, Ben-david, Xu, Chhablani,
  Wang, Fries, Al-shaibani, Sharma, Thakker, Almubarak, Tang, Radev, Jiang, and
  Rush}]{bach-etal-2022-promptsource}
Stephen Bach, Victor Sanh, Zheng~Xin Yong, Albert Webson, Colin Raffel,
  Nihal~V. Nayak, Abheesht Sharma, Taewoon Kim, M~Saiful Bari, Thibault Fevry,
  Zaid Alyafeai, Manan Dey, Andrea Santilli, Zhiqing Sun, Srulik Ben-david,
  Canwen Xu, Gunjan Chhablani, Han Wang, Jason Fries, Maged Al-shaibani, Shanya
  Sharma, Urmish Thakker, Khalid Almubarak, Xiangru Tang, Dragomir Radev, Mike
  Tian-jian Jiang, and Alexander Rush. 2022.
\newblock \href {https://doi.org/10.18653/v1/2022.acl-demo.9}
  {{P}rompt{S}ource: An integrated development environment and repository for
  natural language prompts}.
\newblock In \emph{Proceedings of the 60th Annual Meeting of the Association
  for Computational Linguistics: System Demonstrations}, pages 93--104, Dublin,
  Ireland. Association for Computational Linguistics.

\bibitem[{Bandarkar et~al.(2024)Bandarkar, Liang, Muller, Artetxe, Shukla,
  Husa, Goyal, Krishnan, Zettlemoyer, and
  Khabsa}]{bandarkar-etal-2024-belebele}
Lucas Bandarkar, Davis Liang, Benjamin Muller, Mikel Artetxe, Satya~Narayan
  Shukla, Donald Husa, Naman Goyal, Abhinandan Krishnan, Luke Zettlemoyer, and
  Madian Khabsa. 2024.
\newblock \href {https://doi.org/10.18653/v1/2024.acl-long.44} {The belebele
  benchmark: a parallel reading comprehension dataset in 122 language
  variants}.
\newblock In \emph{Proceedings of the 62nd Annual Meeting of the Association
  for Computational Linguistics (Volume 1: Long Papers)}, pages 749--775,
  Bangkok, Thailand. Association for Computational Linguistics.

\bibitem[{Barbaresi(2021)}]{barbaresi-2021-trafilatura}
Adrien Barbaresi. 2021.
\newblock \href {https://doi.org/10.18653/v1/2021.acl-demo.15} {Trafilatura:
  {A} web scraping library and command-line tool for text discovery and
  extraction}.
\newblock In \emph{Proceedings of the 59th Annual Meeting of the Association
  for Computational Linguistics and the 11th International Joint Conference on
  Natural Language Processing: System Demonstrations}, pages 122--131, Online.
  Association for Computational Linguistics.

\bibitem[{Baucells et~al.(2025)Baucells, Aula-Blasco, de~Dios-Flores,
  Paniagua~Su{\'a}rez, Perez, Salles, Sotelo~Docio, Falc{\~a}o, Saiz,
  Sepulveda~Torres, Barnes, Gamallo, Gonzalez-Agirre, Rigau, and
  Villegas}]{baucells-etal-2025-iberobench}
Irene Baucells, Javier Aula-Blasco, Iria de~Dios-Flores, Silvia
  Paniagua~Su{\'a}rez, Naiara Perez, Anna Salles, Susana Sotelo~Docio,
  J{\'u}lia Falc{\~a}o, Jose~Javier Saiz, Robiert Sepulveda~Torres, Jeremy
  Barnes, Pablo Gamallo, Aitor Gonzalez-Agirre, German Rigau, and Marta
  Villegas. 2025.
\newblock \href {https://aclanthology.org/2025.coling-main.699/}
  {{I}bero{B}ench: A benchmark for {LLM} evaluation in {I}berian languages}.
\newblock In \emph{Proceedings of the 31st International Conference on
  Computational Linguistics}, pages 10491--10519, Abu Dhabi, UAE. Association
  for Computational Linguistics.

\bibitem[{Burchell et~al.(2023)Burchell, Birch, Bogoychev, and
  Heafield}]{burchell-etal-2023-open}
Laurie Burchell, Alexandra Birch, Nikolay Bogoychev, and Kenneth Heafield.
  2023.
\newblock \href {https://doi.org/10.18653/v1/2023.acl-short.75} {An open
  dataset and model for language identification}.
\newblock In \emph{Proceedings of the 61st Annual Meeting of the Association
  for Computational Linguistics (Volume 2: Short Papers)}, pages 865--879,
  Toronto, Canada. Association for Computational Linguistics.

\bibitem[{Burchell et~al.(2025)Burchell, De~Gibert~Bonet, Arefyev, Aulamo,
  Ba{\~n}{\'o}n, Chen, Fedorova, Guillou, Haddow, Haji{\v{c}}, Helcl,
  Henriksson, Klimaszewski, Komulainen, Kutuzov, Kyt{\"o}niemi, Laippala,
  M{\ae}hlum, Malik, Mehryary, Mikhailov, Moghe, Myntti, O{'}Brien, Oepen, Pal,
  Piha, Pyysalo, Ram{\'i}rez-S{\'a}nchez, Samuel, Stepachev, Tiedemann,
  Vari{\v{s}}, Vojt{\v{e}}chov{\'a}, and
  Zaragoza-Bernabeu}]{burchell-etal-2025-expanded}
Laurie Burchell, Ona De~Gibert~Bonet, Nikolay Arefyev, Mikko Aulamo, Marta
  Ba{\~n}{\'o}n, Pinzhen Chen, Mariia Fedorova, Liane Guillou, Barry Haddow,
  Jan Haji{\v{c}}, Jind{\v{r}}ich Helcl, Erik Henriksson, Mateusz Klimaszewski,
  Ville Komulainen, Andrey Kutuzov, Joona Kyt{\"o}niemi, Veronika Laippala,
  Petter M{\ae}hlum, Bhavitvya Malik, Farrokh Mehryary, Vladislav Mikhailov,
  Nikita Moghe, Amanda Myntti, Dayy{\'a}n O{'}Brien, Stephan Oepen, Proyag Pal,
  Jousia Piha, Sampo Pyysalo, Gema Ram{\'i}rez-S{\'a}nchez, David Samuel, Pavel
  Stepachev, J{\"o}rg Tiedemann, Du{\v{s}}an Vari{\v{s}}, Tereza
  Vojt{\v{e}}chov{\'a}, and Jaume Zaragoza-Bernabeu. 2025.
\newblock \href {https://aclanthology.org/2025.acl-long.854/} {An expanded
  massive multilingual dataset for high-performance language technologies
  ({HPLT})}.
\newblock In \emph{Proceedings of the 63rd Annual Meeting of the Association
  for Computational Linguistics (Volume 1: Long Papers)}, pages 17452--17485,
  Vienna, Austria. Association for Computational Linguistics.

\bibitem[{Burchell et~al.(2024)Burchell, Maillard, Anastasopoulos, Federmann,
  Koehn, and Wang}]{burchell-etal-2024-findings}
Laurie Burchell, Jean Maillard, Antonios Anastasopoulos, Christian Federmann,
  Philipp Koehn, and Skyler Wang. 2024.
\newblock \href {https://doi.org/10.18653/v1/2024.wmt-1.4} {Findings of the
  {WMT} 2024 shared task of the open language data initiative}.
\newblock In \emph{Proceedings of the Ninth Conference on Machine Translation},
  pages 110--117, Miami, Florida, USA. Association for Computational
  Linguistics.

\bibitem[{Charpentier and Samuel(2024)}]{charpentier-samuel-2024-bert}
Lucas Georges~Gabriel Charpentier and David Samuel. 2024.
\newblock \href {https://aclanthology.org/2024.conll-babylm.24/} {{GPT} or
  {BERT}: why not both?}
\newblock In \emph{The 2nd BabyLM Challenge at the 28th Conference on
  Computational Natural Language Learning}, pages 262--283, Miami, FL, USA.
  Association for Computational Linguistics.

\bibitem[{Conneau et~al.(2020)Conneau, Khandelwal, Goyal, Chaudhary, Wenzek,
  Guzm{\'a}n, Grave, Ott, Zettlemoyer, and
  Stoyanov}]{conneau-etal-2020-unsupervised}
Alexis Conneau, Kartikay Khandelwal, Naman Goyal, Vishrav Chaudhary, Guillaume
  Wenzek, Francisco Guzm{\'a}n, Edouard Grave, Myle Ott, Luke Zettlemoyer, and
  Veselin Stoyanov. 2020.
\newblock \href {https://doi.org/10.18653/v1/2020.acl-main.747} {Unsupervised
  cross-lingual representation learning at scale}.
\newblock In \emph{Proceedings of the 58th Annual Meeting of the Association
  for Computational Linguistics}, pages 8440--8451, Online. Association for
  Computational Linguistics.

\bibitem[{de~Gibert et~al.(2025)de~Gibert, Attieh, Vahtola, Aulamo, Li,
  V{\'a}zquez, Hu, and Tiedemann}]{degibert2025scaling}
Ona de~Gibert, Joseph Attieh, Teemu Vahtola, Mikko Aulamo, Zihao Li, Ra{\'u}l
  V{\'a}zquez, Tiancheng Hu, and J{\"o}rg Tiedemann. 2025.
\newblock Scaling low-resource mt via synthetic data generation with llms.
\newblock \emph{arXiv preprint arXiv:2505.14423}.

\bibitem[{de~Gibert et~al.(2024)de~Gibert, Nail, Arefyev, Ba{\~n}{\'o}n,
  van~der Linde, Ji, Zaragoza-Bernabeu, Aulamo, Ram{\'i}rez-S{\'a}nchez,
  Kutuzov, Pyysalo, Oepen, and Tiedemann}]{de-gibert-etal-2024-new}
Ona de~Gibert, Graeme Nail, Nikolay Arefyev, Marta Ba{\~n}{\'o}n, Jelmer
  van~der Linde, Shaoxiong Ji, Jaume Zaragoza-Bernabeu, Mikko Aulamo, Gema
  Ram{\'i}rez-S{\'a}nchez, Andrey Kutuzov, Sampo Pyysalo, Stephan Oepen, and
  J{\"o}rg Tiedemann. 2024.
\newblock \href {https://aclanthology.org/2024.lrec-main.100/} {A new massive
  multilingual dataset for high-performance language technologies}.
\newblock In \emph{Proceedings of the 2024 Joint International Conference on
  Computational Linguistics, Language Resources and Evaluation (LREC-COLING
  2024)}, pages 1116--1128, Torino, Italia. ELRA and ICCL.

\bibitem[{Dementieva et~al.(2024)Dementieva, Khylenko, Babakov, and
  Groh}]{dementieva-etal-2024-toxicity}
Daryna Dementieva, Valeriia Khylenko, Nikolay Babakov, and Georg Groh. 2024.
\newblock \href {https://doi.org/10.18653/v1/2024.woah-1.19} {Toxicity
  classification in {U}krainian}.
\newblock In \emph{Proceedings of the 8th Workshop on Online Abuse and Harms
  (WOAH 2024)}, pages 244--255, Mexico City, Mexico. Association for
  Computational Linguistics.

\bibitem[{Deutsch et~al.(2025{\natexlab{a}})Deutsch, Briakou, Caswell,
  Finkelstein, Galor, Juraska, Kovacs, Lui, Rei, Riesa
  et~al.}]{deutsch2025wmt24++}
Daniel Deutsch, Eleftheria Briakou, Isaac Caswell, Mara Finkelstein, Rebecca
  Galor, Juraj Juraska, Geza Kovacs, Alison Lui, Ricardo Rei, Jason Riesa,
  et~al. 2025{\natexlab{a}}.
\newblock Wmt24++: Expanding the language coverage of wmt24 to 55 languages \&
  dialects.
\newblock \emph{arXiv preprint arXiv:2502.12404}.

\bibitem[{Deutsch et~al.(2025{\natexlab{b}})Deutsch, Briakou, Caswell,
  Finkelstein, Galor, Juraska, Kovacs, Lui, Rei, Riesa, Rijhwani, Riley,
  Salesky, Trabelsi, Winkler, Zhang, and Freitag}]{deutsch-etal-2025-wmt24}
Daniel Deutsch, Eleftheria Briakou, Isaac~Rayburn Caswell, Mara Finkelstein,
  Rebecca Galor, Juraj Juraska, Geza Kovacs, Alison Lui, Ricardo Rei, Jason
  Riesa, Shruti Rijhwani, Parker Riley, Elizabeth Salesky, Firas Trabelsi,
  Stephanie Winkler, Biao Zhang, and Markus Freitag. 2025{\natexlab{b}}.
\newblock \href {https://aclanthology.org/2025.findings-acl.634/} {{WMT}24++:
  Expanding the language coverage of {WMT}24 to 55 languages {\&} dialects}.
\newblock In \emph{Findings of the Association for Computational Linguistics:
  ACL 2025}, pages 12257--12284, Vienna, Austria. Association for Computational
  Linguistics.

\bibitem[{Devlin et~al.(2019)Devlin, Chang, Lee, and
  Toutanova}]{devlin-etal-2019-bert}
Jacob Devlin, Ming-Wei Chang, Kenton Lee, and Kristina Toutanova. 2019.
\newblock \href {https://doi.org/10.18653/v1/N19-1423} {{BERT}: Pre-training of
  deep bidirectional transformers for language understanding}.
\newblock In \emph{Proceedings of the 2019 Conference of the North {A}merican
  Chapter of the Association for Computational Linguistics: Human Language
  Technologies, Volume 1 (Long and Short Papers)}, pages 4171--4186,
  Minneapolis, Minnesota. Association for Computational Linguistics.

\bibitem[{Doshi et~al.(2024)Doshi, Dabre, and
  Bhattacharyya}]{doshi-etal-2024-pretraining}
Meet Doshi, Raj Dabre, and Pushpak Bhattacharyya. 2024.
\newblock \href {https://doi.org/10.18653/v1/2024.emnlp-main.334} {Pretraining
  language models using translationese}.
\newblock In \emph{Proceedings of the 2024 Conference on Empirical Methods in
  Natural Language Processing}, pages 5843--5862, Miami, Florida, USA.
  Association for Computational Linguistics.

\bibitem[{Fajcik et~al.(2025)Fajcik, Docekal, Dolezal, Ondrej, Bene{\v{s}},
  Kapsa, Smrz, Polok, Hradis, Neverilova et~al.}]{fajcik2025benczechmark}
Martin Fajcik, Martin Docekal, Jan Dolezal, Karel Ondrej, Karel Bene{\v{s}},
  Jan Kapsa, Pavel Smrz, Alexander Polok, Michal Hradis, Zuzana Neverilova,
  et~al. 2025.
\newblock {BenCzechMark: A Czech-centric Multitask and Multimetric Benchmark
  for Large Language Models with Duel Scoring Mechanism}.
\newblock \emph{Transactions of the Association for Computational Linguistics},
  13:1068--1095.

\bibitem[{Faysse et~al.(2024)Faysse, Fernandes, Guerreiro, Loison, Alves,
  Corro, Boizard, Alves, Rei, Martins et~al.}]{faysse2024croissantllm}
Manuel Faysse, Patrick Fernandes, Nuno~M Guerreiro, Ant{\'o}nio Loison,
  Duarte~Miguel Alves, Caio Corro, Nicolas Boizard, Jo{\~a}o Alves, Ricardo
  Rei, Pedro~Henrique Martins, et~al. 2024.
\newblock {CroissantLLM: A Truly Bilingual French-English Language Model}.
\newblock \emph{Transactions on Machine Learning Research}.

\bibitem[{Fernandes et~al.(2023)Fernandes, Yin, Liu, Martins, and
  Neubig}]{fernandes-etal-2023-translation}
Patrick Fernandes, Kayo Yin, Emmy Liu, Andr{\'e} Martins, and Graham Neubig.
  2023.
\newblock \href {https://doi.org/10.18653/v1/2023.acl-long.36} {When does
  translation require context? a data-driven, multilingual exploration}.
\newblock In \emph{Proceedings of the 61st Annual Meeting of the Association
  for Computational Linguistics (Volume 1: Long Papers)}, pages 606--626,
  Toronto, Canada. Association for Computational Linguistics.

\bibitem[{Fourrier et~al.(2024)Fourrier, Habib, Lozovskaya, Szafer, and
  Wolf}]{fourrier2024open}
Cl{\'e}mentine Fourrier, Nathan Habib, Alina Lozovskaya, Konrad Szafer, and
  Thomas Wolf. 2024.
\newblock Open llm leaderboard v2.

\bibitem[{Gao et~al.(2024)Gao, Tow, Abbasi, Biderman, Black, DiPofi, Foster,
  Golding, Hsu, Le~Noac'h, Li, McDonell, Muennighoff, Ociepa, Phang, Reynolds,
  Schoelkopf, Skowron, Sutawika, Tang, Thite, Wang, Wang, and
  Zou}]{eval-harness}
Leo Gao, Jonathan Tow, Baber Abbasi, Stella Biderman, Sid Black, Anthony
  DiPofi, Charles Foster, Laurence Golding, Jeffrey Hsu, Alain Le~Noac'h,
  Haonan Li, Kyle McDonell, Niklas Muennighoff, Chris Ociepa, Jason Phang,
  Laria Reynolds, Hailey Schoelkopf, Aviya Skowron, Lintang Sutawika, Eric
  Tang, Anish Thite, Ben Wang, Kevin Wang, and Andy Zou. 2024.
\newblock \href {https://doi.org/10.5281/zenodo.12608602} {The language model
  evaluation harness}.

\bibitem[{Henriksson et~al.(2026)Henriksson, Myntti, Hellstr{\"o}m, Eskelinen,
  {Erten-Johansson}, and Laippala}]{Henriksson.etal.2024}
Erik Henriksson, Amanda Myntti, Saara Hellstr{\"o}m, Anni Eskelinen, Selcen
  {Erten-Johansson}, and Veronika Laippala. 2026.
\newblock \href {https://doi.org/10.48550/arXiv.2406.19892} {Automatic register
  identification for the open web using multilingual deep learning}.
\newblock \emph{Natural Language Processing}.

\bibitem[{Ivanyuk-Skulskiy et~al.(2021)Ivanyuk-Skulskiy, Zaliznyi, Reshetar,
  Protsyk, Romanchuk, and Shpihanovych}]{ua_datasets_2021}
Bogdan Ivanyuk-Skulskiy, Anton Zaliznyi, Oleksand Reshetar, Oleksiy Protsyk,
  Bohdan Romanchuk, and Vladyslav Shpihanovych. 2021.
\newblock \href {https://github.com/fido-ai/ua-datasets} {ua\_datasets: a
  collection of ukrainian language datasets}.

\bibitem[{Jentoft and Samuel(2023)}]{jentoft-samuel-2023-nocola}
Matias Jentoft and David Samuel. 2023.
\newblock \href {https://aclanthology.org/2023.nodalida-1.60/} {{N}o{C}o{LA}:
  The {N}orwegian corpus of linguistic acceptability}.
\newblock In \emph{Proceedings of the 24th Nordic Conference on Computational
  Linguistics (NoDaLiDa)}, pages 610--617, T{\'o}rshavn, Faroe Islands.
  University of Tartu Library.

\bibitem[{Jumelet et~al.(2025)Jumelet, Weissweiler, Nivre, and
  Bisazza}]{multiblimp}
Jaap Jumelet, Leonie Weissweiler, Joakim Nivre, and Arianna Bisazza. 2025.
\newblock \href {http://arxiv.org/abs/2504.02768} {Multiblimp 1.0: A massively
  multilingual benchmark of linguistic minimal pairs}.

\bibitem[{Junczys-Dowmunt et~al.(2018)Junczys-Dowmunt, Grundkiewicz, Dwojak,
  Hoang, Heafield, Neckermann, Seide, Germann, Fikri~Aji, Bogoychev, Martins,
  and Birch}]{mariannmt}
Marcin Junczys-Dowmunt, Roman Grundkiewicz, Tomasz Dwojak, Hieu Hoang, Kenneth
  Heafield, Tom Neckermann, Frank Seide, Ulrich Germann, Alham Fikri~Aji,
  Nikolay Bogoychev, Andr\'{e} F.~T. Martins, and Alexandra Birch. 2018.
\newblock \href {http://www.aclweb.org/anthology/P18-4020} {Marian: Fast neural
  machine translation in {C++}}.
\newblock In \emph{Proceedings of ACL 2018, System Demonstrations}, pages
  116--121, Melbourne, Australia. Association for Computational Linguistics.

\bibitem[{Kauf and Ivanova(2023)}]{kauf-ivanova-2023-better}
Carina Kauf and Anna Ivanova. 2023.
\newblock \href {https://doi.org/10.18653/v1/2023.acl-short.80} {A better way
  to do masked language model scoring}.
\newblock In \emph{Proceedings of the 61st Annual Meeting of the Association
  for Computational Linguistics (Volume 2: Short Papers)}, pages 925--935,
  Toronto, Canada. Association for Computational Linguistics.

\bibitem[{Kudugunta et~al.(2023)Kudugunta, Caswell, Zhang, Garcia, Xin,
  Kusupati, Stella, Bapna, and Firat}]{10.5555/3666122.3669062}
Sneha Kudugunta, Isaac Caswell, Biao Zhang, Xavier Garcia, Derrick Xin, Aditya
  Kusupati, Romi Stella, Ankur Bapna, and Orhan Firat. 2023.
\newblock {MADLAD}-400: {A} multilingual and document-level large audited
  dataset.
\newblock In \emph{Proceedings of the 37th International Conference on Neural
  Information Processing Systems}, NeurIPS '23, Red Hook, NY, USA. Curran
  Associates Inc.

\bibitem[{Lee et~al.(2022)Lee, Ippolito, Nystrom, Zhang, Eck, Callison-Burch,
  and Carlini}]{lee-etal-2022-deduplicating}
Katherine Lee, Daphne Ippolito, Andrew Nystrom, Chiyuan Zhang, Douglas Eck,
  Chris Callison-Burch, and Nicholas Carlini. 2022.
\newblock \href {https://doi.org/10.18653/v1/2022.acl-long.577} {Deduplicating
  training data makes language models better}.
\newblock In \emph{Proceedings of the 60th Annual Meeting of the Association
  for Computational Linguistics (Volume 1: Long Papers)}, pages 8424--8445,
  Dublin, Ireland. Association for Computational Linguistics.

\bibitem[{Lozhkov et~al.(2024)Lozhkov, Ben~Allal, von Werra, and
  Wolf}]{lozhkov2024fineweb-edu}
Anton Lozhkov, Loubna Ben~Allal, Leandro von Werra, and Thomas Wolf. 2024.
\newblock \href {https://doi.org/10.57967/hf/2497} {Fineweb-edu: the finest
  collection of educational content}.

\bibitem[{Luukkonen et~al.(2023)Luukkonen, Komulainen, Luoma, Eskelinen,
  Kanerva, Kupari, Ginter, Laippala, Muennighoff, Piktus, Wang, Tazi, Scao,
  Wolf, Suominen, Sairanen, Merioksa, Heinonen, Vahtola, Antao, and
  Pyysalo}]{luukkonen-etal-2023-fingpt}
Risto Luukkonen, Ville Komulainen, Jouni Luoma, Anni Eskelinen, Jenna Kanerva,
  Hanna-Mari Kupari, Filip Ginter, Veronika Laippala, Niklas Muennighoff,
  Aleksandra Piktus, Thomas Wang, Nouamane Tazi, Teven Scao, Thomas Wolf, Osma
  Suominen, Samuli Sairanen, Mikko Merioksa, Jyrki Heinonen, Aija Vahtola,
  Samuel Antao, and Sampo Pyysalo. 2023.
\newblock \href {https://doi.org/10.18653/v1/2023.emnlp-main.164} {{F}in{GPT}:
  Large generative models for a small language}.
\newblock In \emph{Proceedings of the 2023 Conference on Empirical Methods in
  Natural Language Processing}, pages 2710--2726, Singapore. Association for
  Computational Linguistics.

\bibitem[{Marone et~al.(2025)Marone, Weller, Fleshman, Yang, Lawrie, and
  Durme}]{marone2025mmbertmodernmultilingualencoder}
Marc Marone, Orion Weller, William Fleshman, Eugene Yang, Dawn Lawrie, and
  Benjamin~Van Durme. 2025.
\newblock \href {http://arxiv.org/abs/2509.06888} {mmbert: A modern
  multilingual encoder with annealed language learning}.

\bibitem[{Merx et~al.(2025)Merx, Suominen, Cohn, and
  Vylomova}]{merx2025openwho}
Rapha{\"e}l Merx, Hanna Suominen, Trevor Cohn, and Ekaterina Vylomova. 2025.
\newblock Openwho: A document-level parallel corpus for health translation in
  low-resource languages.
\newblock \emph{arXiv preprint arXiv:2508.16048}.

\bibitem[{Mikhailov et~al.(2025)Mikhailov, Enstad, Samuel, Farseth{\r{a}}s,
  Kutuzov, Velldal, and {\O}vrelid}]{mikhailov-etal-2025-noreval}
Vladislav Mikhailov, Tita Enstad, David Samuel, Hans~Christian Farseth{\r{a}}s,
  Andrey Kutuzov, Erik Velldal, and Lilja {\O}vrelid. 2025.
\newblock \href {https://aclanthology.org/2025.findings-acl.181/} {{N}or{E}val:
  A {N}orwegian language understanding and generation evaluation benchmark}.
\newblock In \emph{Findings of the Association for Computational Linguistics:
  ACL 2025}, pages 3495--3541, Vienna, Austria. Association for Computational
  Linguistics.

\bibitem[{Misra(2022)}]{misra2022minicons}
Kanishka Misra. 2022.
\newblock minicons: Enabling flexible behavioral and representational analyses
  of transformer language models.
\newblock \emph{arXiv preprint arXiv:2203.13112}.

\bibitem[{Muennighoff et~al.(2023{\natexlab{a}})Muennighoff, Rush, Barak, Scao,
  Tazi, Piktus, Pyysalo, Wolf, and Raffel}]{muennighoff2023scaling}
Niklas Muennighoff, Alexander~M Rush, Boaz Barak, Teven~Le Scao, Nouamane Tazi,
  Aleksandra Piktus, Sampo Pyysalo, Thomas Wolf, and Colin Raffel.
  2023{\natexlab{a}}.
\newblock \href {https://openreview.net/forum?id=j5BuTrEj35} {{Scaling
  Data-Constrained Language Models}}.
\newblock In \emph{Thirty-seventh Conference on Neural Information Processing
  Systems}.

\bibitem[{Muennighoff et~al.(2023{\natexlab{b}})Muennighoff, Wang, Sutawika,
  Roberts, Biderman, Le~Scao, Bari, Shen, Yong, Schoelkopf, Tang, Radev, Aji,
  Almubarak, Albanie, Alyafeai, Webson, Raff, and
  Raffel}]{muennighoff-etal-2023-crosslingual}
Niklas Muennighoff, Thomas Wang, Lintang Sutawika, Adam Roberts, Stella
  Biderman, Teven Le~Scao, M~Saiful Bari, Sheng Shen, Zheng~Xin Yong, Hailey
  Schoelkopf, Xiangru Tang, Dragomir Radev, Alham~Fikri Aji, Khalid Almubarak,
  Samuel Albanie, Zaid Alyafeai, Albert Webson, Edward Raff, and Colin Raffel.
  2023{\natexlab{b}}.
\newblock \href {https://doi.org/10.18653/v1/2023.acl-long.891} {Crosslingual
  generalization through multitask finetuning}.
\newblock In \emph{Proceedings of the 61st Annual Meeting of the Association
  for Computational Linguistics (Volume 1: Long Papers)}, pages 15991--16111,
  Toronto, Canada. Association for Computational Linguistics.

\bibitem[{O'Brien et~al.(2025)O'Brien, Malik, de~Gibert, Chen, Haddow, and
  Tiedemann}]{obrien2025dochplt}
Dayy{\'a}n O'Brien, Bhavitvya Malik, Ona de~Gibert, Pinzhen Chen, Barry Haddow,
  and J{\"o}rg Tiedemann. 2025.
\newblock Dochplt: A massively multilingual document-level translation dataset.
\newblock \emph{arXiv preprint arXiv:2508.13079}.

\bibitem[{Penedo et~al.(2024)Penedo, Kydl{\'\i}{\v{c}}ek, Lozhkov, Mitchell,
  Raffel, Von~Werra, Wolf et~al.}]{penedo2024fineweb}
Guilherme Penedo, Hynek Kydl{\'\i}{\v{c}}ek, Anton Lozhkov, Margaret Mitchell,
  Colin~A Raffel, Leandro Von~Werra, Thomas Wolf, et~al. 2024.
\newblock The {F}ine{W}eb datasets: {D}ecanting the web for the finest text
  data at scale.
\newblock \emph{Advances in Neural Information Processing Systems},
  37:30811--30849.

\bibitem[{Penedo et~al.(2025)Penedo, Kydlíček, Sabolčec, Messmer, Foroutan,
  Kargaran, Raffel, Jaggi, Werra, and Wolf}]{penedo2025fineweb2pipelinescale}
Guilherme Penedo, Hynek Kydlíček, Vinko Sabolčec, Bettina Messmer, Negar
  Foroutan, Amir~Hossein Kargaran, Colin Raffel, Martin Jaggi, Leandro~Von
  Werra, and Thomas Wolf. 2025.
\newblock \href {http://arxiv.org/abs/2506.20920} {{F}ine{W}eb2: {O}ne pipeline
  to scale them all -- adapting pre-training data processing to every
  language}.

\bibitem[{Pezeshkpour and Hruschka(2024)}]{pezeshkpour-hruschka-2024-large}
Pouya Pezeshkpour and Estevam Hruschka. 2024.
\newblock \href {https://doi.org/10.18653/v1/2024.findings-naacl.130} {Large
  language models sensitivity to the order of options in multiple-choice
  questions}.
\newblock In \emph{Findings of the Association for Computational Linguistics:
  NAACL 2024}, pages 2006--2017, Mexico City, Mexico. Association for
  Computational Linguistics.

\bibitem[{Qorib et~al.(2025)Qorib, Li, and Ng}]{qorib-etal-2025-just}
Muhammad~Reza Qorib, Junyi Li, and Hwee~Tou Ng. 2025.
\newblock \href {https://aclanthology.org/2025.acl-long.1602/} {Just go
  parallel: Improving the multilingual capabilities of large language models}.
\newblock In \emph{Proceedings of the 63rd Annual Meeting of the Association
  for Computational Linguistics (Volume 1: Long Papers)}, pages 33411--33424,
  Vienna, Austria. Association for Computational Linguistics.

\bibitem[{Raffel et~al.(2020)Raffel, Shazeer, Roberts, Lee, Narang, Matena,
  Zhou, Li, and Liu}]{JMLR:v21:20-074}
Colin Raffel, Noam Shazeer, Adam Roberts, Katherine Lee, Sharan Narang, Michael
  Matena, Yanqi Zhou, Wei Li, and Peter~J. Liu. 2020.
\newblock \href {http://jmlr.org/papers/v21/20-074.html} {Exploring the limits
  of transfer learning with a unified text-to-text transformer}.
\newblock \emph{Journal of Machine Learning Research}, 21(140):1--67.

\bibitem[{Rahimi et~al.(2019)Rahimi, Li, and Cohn}]{rahimi-etal-2019-massively}
Afshin Rahimi, Yuan Li, and Trevor Cohn. 2019.
\newblock \href {https://doi.org/10.18653/v1/P19-1015} {Massively multilingual
  transfer for {NER}}.
\newblock In \emph{Proceedings of the 57th Annual Meeting of the Association
  for Computational Linguistics}, pages 151--164, Florence, Italy. Association
  for Computational Linguistics.

\bibitem[{Ramos et~al.(2025)Ramos, Fernandes, Agrawal, and
  Martins}]{ramos2025multilingual}
Miguel~Moura Ramos, Patrick Fernandes, Sweta Agrawal, and Andr{\'e}~FT Martins.
  2025.
\newblock Multilingual contextualization of large language models for
  document-level machine translation.
\newblock \emph{arXiv preprint arXiv:2504.12140}.

\bibitem[{Rofin et~al.(2023)Rofin, Mikhailov, Florinsky, Kravchenko, Shavrina,
  Tutubalina, Karabekyan, and Artemova}]{rofin-etal-2023-votenrank}
Mark Rofin, Vladislav Mikhailov, Mikhail Florinsky, Andrey Kravchenko, Tatiana
  Shavrina, Elena Tutubalina, Daniel Karabekyan, and Ekaterina Artemova. 2023.
\newblock \href {https://doi.org/10.18653/v1/2023.eacl-main.48}
  {Vote{'}n{'}rank: Revision of benchmarking with social choice theory}.
\newblock In \emph{Proceedings of the 17th Conference of the European Chapter
  of the Association for Computational Linguistics}, pages 670--686, Dubrovnik,
  Croatia. Association for Computational Linguistics.

\bibitem[{Romanou et~al.(2025)Romanou, Foroutan, Sotnikova, Nelaturu, Singh,
  Maheshwary, Altomare, Chen, Haggag, A, Amayuelas, Amirudin, Boiko, Chang,
  Chim, Cohen, Dalmia, Diress, Duwal, Dzenhaliou, Florez, Farestam, Imperial,
  Islam, Isotalo, Jabbarishiviari, Karlsson, Khalilov, Klamm, Koto,
  Krzemi{\'n}ski, de~Melo, Montariol, Nan, Niklaus, Novikova, Ceron, Paul,
  Ploeger, Purbey, Rajwal, Ravi, Rydell, Santhosh, Sharma, Skenduli, Moakhar,
  soltani moakhar, Tarun, Wasi, Weerasinghe, Yilmaz, Zhang, Schlag, Fadaee,
  Hooker, and Bosselut}]{romanou2025include}
Angelika Romanou, Negar Foroutan, Anna Sotnikova, Sree~Harsha Nelaturu,
  Shivalika Singh, Rishabh Maheshwary, Micol Altomare, Zeming Chen, Mohamed~A.
  Haggag, Snegha A, Alfonso Amayuelas, Azril~Hafizi Amirudin, Danylo Boiko,
  Michael Chang, Jenny Chim, Gal Cohen, Aditya~Kumar Dalmia, Abraham Diress,
  Sharad Duwal, Daniil Dzenhaliou, Daniel Fernando~Erazo Florez, Fabian
  Farestam, Joseph~Marvin Imperial, Shayekh~Bin Islam, Perttu Isotalo, Maral
  Jabbarishiviari, B{\"o}rje~F. Karlsson, Eldar Khalilov, Christopher Klamm,
  Fajri Koto, Dominik Krzemi{\'n}ski, Gabriel~Adriano de~Melo, Syrielle
  Montariol, Yiyang Nan, Joel Niklaus, Jekaterina Novikova, Johan Samir~Obando
  Ceron, Debjit Paul, Esther Ploeger, Jebish Purbey, Swati Rajwal,
  Selvan~Sunitha Ravi, Sara Rydell, Roshan Santhosh, Drishti Sharma,
  Marjana~Prifti Skenduli, Arshia~Soltani Moakhar, Bardia soltani moakhar,
  Ayush~Kumar Tarun, Azmine~Toushik Wasi, Thenuka~Ovin Weerasinghe, Serhan
  Yilmaz, Mike Zhang, Imanol Schlag, Marzieh Fadaee, Sara Hooker, and Antoine
  Bosselut. 2025.
\newblock \href {https://openreview.net/forum?id=k3gCieTXeY} {{INCLUDE}:
  Evaluating multilingual language understanding with regional knowledge}.
\newblock In \emph{The Thirteenth International Conference on Learning
  Representations}.

\bibitem[{Romanyshyn et~al.(2024)Romanyshyn, Syvokon, and
  Kyslyi}]{romanyshyn-etal-2024-unlp}
Mariana Romanyshyn, Oleksiy Syvokon, and Roman Kyslyi. 2024.
\newblock \href {https://aclanthology.org/2024.unlp-1.9/} {The {UNLP} 2024
  shared task on fine-tuning large language models for {U}krainian}.
\newblock In \emph{Proceedings of the Third Ukrainian Natural Language
  Processing Workshop (UNLP) @ LREC-COLING 2024}, pages 67--74, Torino, Italia.
  ELRA and ICCL.

\bibitem[{Samuel et~al.(2023)Samuel, Kutuzov, Touileb, Velldal, {\O}vrelid,
  R{\o}nningstad, Sigdel, and Palatkina}]{samuel-etal-2023-norbench}
David Samuel, Andrey Kutuzov, Samia Touileb, Erik Velldal, Lilja {\O}vrelid,
  Egil R{\o}nningstad, Elina Sigdel, and Anna Palatkina. 2023.
\newblock \href {https://aclanthology.org/2023.nodalida-1.61/} {{N}or{B}ench
  {--} a benchmark for {N}orwegian language models}.
\newblock In \emph{Proceedings of the 24th Nordic Conference on Computational
  Linguistics (NoDaLiDa)}, pages 618--633, T{\'o}rshavn, Faroe Islands.
  University of Tartu Library.

\bibitem[{Sclar et~al.(2024)Sclar, Choi, Tsvetkov, and
  Suhr}]{sclar2024quantifying}
Melanie Sclar, Yejin Choi, Yulia Tsvetkov, and Alane Suhr. 2024.
\newblock \href {https://openreview.net/forum?id=RIu5lyNXjT} {{Quantifying
  Language Models' Sensitivity to Spurious Features in Prompt Design or: How I
  learned to start worrying about prompt formatting}}.
\newblock In \emph{The Twelfth International Conference on Learning
  Representations}.

\bibitem[{Shoeybi et~al.(2019)Shoeybi, Patwary, Puri, LeGresley, Casper, and
  Catanzaro}]{megatron-lm}
Mohammad Shoeybi, Mostofa Patwary, Raul Puri, Patrick LeGresley, Jared Casper,
  and Bryan Catanzaro. 2019.
\newblock {Megatron-LM: Training Multi-Billion Parameter Language Models Using
  Model Parallelism}.
\newblock \emph{arXiv preprint arXiv:1909.08053}.

\bibitem[{Singh et~al.(2025)Singh, Romanou, Fourrier, Adelani, Ngui,
  Vila-Suero, Limkonchotiwat, Marchisio, Leong, Susanto, Ng, Longpre, Ruder,
  Ko, Bosselut, Oh, Martins, Choshen, Ippolito, Ferrante, Fadaee, Ermis, and
  Hooker}]{singh-etal-2025-global}
Shivalika Singh, Angelika Romanou, Cl{\'e}mentine Fourrier, David~Ifeoluwa
  Adelani, Jian~Gang Ngui, Daniel Vila-Suero, Peerat Limkonchotiwat, Kelly
  Marchisio, Wei~Qi Leong, Yosephine Susanto, Raymond Ng, Shayne Longpre,
  Sebastian Ruder, Wei-Yin Ko, Antoine Bosselut, Alice Oh, Andre Martins,
  Leshem Choshen, Daphne Ippolito, Enzo Ferrante, Marzieh Fadaee, Beyza Ermis,
  and Sara Hooker. 2025.
\newblock \href {https://aclanthology.org/2025.acl-long.919/} {Global {MMLU}:
  Understanding and addressing cultural and linguistic biases in multilingual
  evaluation}.
\newblock In \emph{Proceedings of the 63rd Annual Meeting of the Association
  for Computational Linguistics (Volume 1: Long Papers)}, pages 18761--18799,
  Vienna, Austria. Association for Computational Linguistics.

\bibitem[{Su et~al.(2025)Su, Kong, Lin, Jennings, Norick, Kliegl, Patwary,
  Shoeybi, and Catanzaro}]{su-etal-2025-nemotron}
Dan Su, Kezhi Kong, Ying Lin, Joseph Jennings, Brandon Norick, Markus Kliegl,
  Mostofa Patwary, Mohammad Shoeybi, and Bryan Catanzaro. 2025.
\newblock \href {https://aclanthology.org/2025.acl-long.123/} {Nemotron-{CC}:
  Transforming {C}ommon {C}rawl into a refined long-horizon pretraining
  dataset}.
\newblock In \emph{Proceedings of the 63rd Annual Meeting of the Association
  for Computational Linguistics (Volume 1: Long Papers)}, pages 2459--2475,
  Vienna, Austria. Association for Computational Linguistics.

\bibitem[{Tay et~al.(2023)Tay, Dehghani, Tran, Garcia, Wei, Wang, Chung, Bahri,
  Schuster, Zheng, Zhou, Houlsby, and Metzler}]{tay2023ul}
Yi~Tay, Mostafa Dehghani, Vinh~Q. Tran, Xavier Garcia, Jason Wei, Xuezhi Wang,
  Hyung~Won Chung, Dara Bahri, Tal Schuster, Steven Zheng, Denny Zhou, Neil
  Houlsby, and Donald Metzler. 2023.
\newblock \href {https://openreview.net/forum?id=6ruVLB727MC} {{UL}2: Unifying
  language learning paradigms}.
\newblock In \emph{The Eleventh International Conference on Learning
  Representations}.

\bibitem[{Team(2025)}]{Gemma-3}
Gemma Team. 2025.
\newblock \href {https://goo.gle/Gemma3Report} {Gemma 3}.
\newblock Technical report, Google.

\bibitem[{Team et~al.(2022)Team, Costa-jussà, Cross, Çelebi, Elbayad,
  Heafield, Heffernan, Kalbassi, Lam, Licht, Maillard, Sun, Wang, Wenzek,
  Youngblood, Akula, Barrault, Gonzalez, Hansanti, Hoffman, Jarrett, Sadagopan,
  Rowe, Spruit, Tran, Andrews, Ayan, Bhosale, Edunov, Fan, Gao, Goswami,
  Guzmán, Koehn, Mourachko, Ropers, Saleem, Schwenk, and
  Wang}]{nllbteam2022languageleftbehindscaling}
NLLB Team, Marta~R. Costa-jussà, James Cross, Onur Çelebi, Maha Elbayad,
  Kenneth Heafield, Kevin Heffernan, Elahe Kalbassi, Janice Lam, Daniel Licht,
  Jean Maillard, Anna Sun, Skyler Wang, Guillaume Wenzek, Al~Youngblood, Bapi
  Akula, Loic Barrault, Gabriel~Mejia Gonzalez, Prangthip Hansanti, John
  Hoffman, Semarley Jarrett, Kaushik~Ram Sadagopan, Dirk Rowe, Shannon Spruit,
  Chau Tran, Pierre Andrews, Necip~Fazil Ayan, Shruti Bhosale, Sergey Edunov,
  Angela Fan, Cynthia Gao, Vedanuj Goswami, Francisco Guzmán, Philipp Koehn,
  Alexandre Mourachko, Christophe Ropers, Safiyyah Saleem, Holger Schwenk, and
  Jeff Wang. 2022.
\newblock \href {http://arxiv.org/abs/2207.04672} {No language left behind:
  Scaling human-centered machine translation}.

\bibitem[{Tiedemann et~al.(2023)Tiedemann, Aulamo, Bakshandaeva, Boggia,
  Gr{\"o}nroos, Nieminen, Raganato, Scherrer, Vazquez, and
  Virpioja}]{tiedemann2023democratizing}
J{\"o}rg Tiedemann, Mikko Aulamo, Daria Bakshandaeva, Michele Boggia, Stig-Arne
  Gr{\"o}nroos, Tommi Nieminen, Alessandro Raganato, Yves Scherrer, Raul
  Vazquez, and Sami Virpioja. 2023.
\newblock Democratizing neural machine translation with {OPUS-MT}.
\newblock \emph{Language Resources and Evaluation}, pages 1--43.

\bibitem[{Tiedemann and de~Gibert(2023)}]{tiedemann-de-gibert-2023-opus}
J{\"o}rg Tiedemann and Ona de~Gibert. 2023.
\newblock \href {https://doi.org/10.18653/v1/2023.acl-demo.30} {The {OPUS}-{MT}
  dashboard {--} a toolkit for a systematic evaluation of open machine
  translation models}.
\newblock In \emph{Proceedings of the 61st Annual Meeting of the Association
  for Computational Linguistics (Volume 3: System Demonstrations)}, pages
  315--327, Toronto, Canada. Association for Computational Linguistics.

\bibitem[{Touvron et~al.(2023)Touvron, Lavril, Izacard, Martinet, Lachaux,
  Lacroix, Rozi{\`e}re, Goyal, Hambro, Azhar et~al.}]{touvron2023llama}
Hugo Touvron, Thibaut Lavril, Gautier Izacard, Xavier Martinet, Marie-Anne
  Lachaux, Timoth{\'e}e Lacroix, Baptiste Rozi{\`e}re, Naman Goyal, Eric
  Hambro, Faisal Azhar, et~al. 2023.
\newblock {Llama: Open and efficient foundation language models}.
\newblock \emph{arXiv preprint arXiv:2302.13971}.

\bibitem[{{\"U}st{\"u}n et~al.(2024){\"U}st{\"u}n, Aryabumi, Yong, Ko,
  D{'}souza, Onilude, Bhandari, Singh, Ooi, Kayid, Vargus, Blunsom, Longpre,
  Muennighoff, Fadaee, Kreutzer, and Hooker}]{ustun-etal-2024-aya}
Ahmet {\"U}st{\"u}n, Viraat Aryabumi, Zheng Yong, Wei-Yin Ko, Daniel D{'}souza,
  Gbemileke Onilude, Neel Bhandari, Shivalika Singh, Hui-Lee Ooi, Amr Kayid,
  Freddie Vargus, Phil Blunsom, Shayne Longpre, Niklas Muennighoff, Marzieh
  Fadaee, Julia Kreutzer, and Sara Hooker. 2024.
\newblock \href {https://doi.org/10.18653/v1/2024.acl-long.845} {Aya model: An
  instruction finetuned open-access multilingual language model}.
\newblock In \emph{Proceedings of the 62nd Annual Meeting of the Association
  for Computational Linguistics (Volume 1: Long Papers)}, pages 15894--15939,
  Bangkok, Thailand. Association for Computational Linguistics.

\bibitem[{Wang et~al.(2025)Wang, Lu, Weber, Ryabinin, Adelani, Chen, Tang, and
  Stenetorp}]{wang2025multilinguallanguagemodelpretraining}
Jiayi Wang, Yao Lu, Maurice Weber, Max Ryabinin, David Adelani, Yihong Chen,
  Raphael Tang, and Pontus Stenetorp. 2025.
\newblock \href {http://arxiv.org/abs/2502.13252} {Multilingual language model
  pretraining using machine-translated data}.

\bibitem[{Xue et~al.(2021)Xue, Constant, Roberts, Kale, Al-Rfou, Siddhant,
  Barua, and Raffel}]{xue-etal-2021-mt5}
Linting Xue, Noah Constant, Adam Roberts, Mihir Kale, Rami Al-Rfou, Aditya
  Siddhant, Aditya Barua, and Colin Raffel. 2021.
\newblock \href {https://doi.org/10.18653/v1/2021.naacl-main.41} {m{T}5: A
  massively multilingual pre-trained text-to-text transformer}.
\newblock In \emph{Proceedings of the 2021 Conference of the North American
  Chapter of the Association for Computational Linguistics: Human Language
  Technologies}, pages 483--498, Online. Association for Computational
  Linguistics.

\bibitem[{Zhang et~al.(2025)Zhang, Moiseev, Ainslie, Suganthan, Ma,
  Bhupatiraju, Lebron, Firat, Joulin, and
  Dong}]{zhang2025encoderdecodergemmaimprovingqualityefficiency}
Biao Zhang, Fedor Moiseev, Joshua Ainslie, Paul Suganthan, Min Ma, Surya
  Bhupatiraju, Fede Lebron, Orhan Firat, Armand Joulin, and Zhe Dong. 2025.
\newblock \href {http://arxiv.org/abs/2504.06225} {Encoder-decoder gemma:
  Improving the quality-efficiency trade-off via adaptation}.

\bibitem[{Zhang et~al.(2024)Zhang, Wang, Liu, Wang, Wang, Li, Sun, and
  Liu}]{zhang2024enhancing}
Yuanchi Zhang, Yile Wang, Zijun Liu, Shuo Wang, Xiaolong Wang, Peng Li, Maosong
  Sun, and Yang Liu. 2024.
\newblock Enhancing multilingual capabilities of large language models through
  self-distillation from resource-rich languages.
\newblock In \emph{Proceedings of the 62nd Annual Meeting of the Association
  for Computational Linguistics (Volume 1: Long Papers)}, pages 11189--11204.

\end{thebibliography}

\newpage
\appendix
\section{Evaluation of monolingual encoder and encoder--decoder models}
\label{ax:models}
T5 training happened before language identification manual inspection, which revealed issues with Asturian, Bosnian, and Croatian. Thus, we did not train GPT-BERT on these languages. A significant part of Asturian corpus turned out to be in fact Spanish (according to reports from native speakers). Serbian in Latin script class was missing from HPLT 3.0 language identifier, which caused documents in this language to be mostly classified as Bosnian, and sometimes also Croatian. We leave training GPT-BERTs on Bosnian, Croatian, and Serbian in Latin script for future work.

\subsection{MultiBLIMP}
MultiBLIMP does not differentiate Bosnian and Croatian in Latin script, thus we used the same `hbs' subset to evaluate Bosnian and Croatian T5 models. 

MultiBLIMP contains two Albanian varieties, Tosk and Gneg one. Our T5 model was only evaluated for the Tosk one.

MultiBLIMP lacks Chinese, Japanese, Indonesian, Korean, Luxembourgish, Norwegian (we use NoCOLA to compensate for that), Serbian in Cyrillic script, Swahili, Thai, Vietnamese.

\subsection{Named Entity Recognition}


In HPLT 2.0, Bosnian NER was tested on Croatian. We tested \texttt{google/mt5-base} on Bosnian.

On average, \HPLT\ T5 models achieve the same performance as HPLT BERTs on the NER task, while at the same time possessing all the advantages of encoder--decoder models in comparison to encoder-only architectures; they also outperform both mT5 models on the MultiBLIMP task.

\subsection{Universal Dependencies}

We use UD version 2.13 for most languages, to ensure comparability with HPLT 1.2 and HPLT 2.0 models. For \texttt{kat\_Geor} (Georgian), we use version 2.15, for \texttt{tha\_Thai} (Thai) and \texttt{ara\_Arab} (Arabic) we use version 2.17. For Albanian, Luxembourgish, Northern Kurdish, Serbian and Tatar, UD lacks datasets or they are too small to meaningfully fine-tune our models, so we skip these languages.

We release the fine-tuned UD parsers at \url{https://huggingface.co/collections/HPLT/ud-parsers}.




\begin{table*}[!htp]
\centering
\small
\adjustbox{max width=\linewidth}{
\begin{tabular}{l|cccccc|cccccc|cccccc|cccccc}
\toprule
\textbf{} &\multicolumn{6}{c}{\textbf{POS Tags}} &\multicolumn{6}{c}{\textbf{Lemmas}} &\multicolumn{6}{c}{\textbf{Dependency Parsing}} &\multicolumn{6}{c}{\textbf{Named Entity Recognition}} \\
\multirow{2}{*}{\textbf{Language}} & \multirow{2}{*}{I} & \multirow{2}{*}{II} & \multirow{2}{*}{III} & {HPLT} & {HPLT} & {HPLT} & \multirow{2}{*}{I} & \multirow{2}{*}{II} & \multirow{2}{*}{III} & {HPLT} & {HPLT} & {HPLT} & \multirow{2}{*}{I} & \multirow{2}{*}{II} & \multirow{2}{*}{III} & {HPLT} & {HPLT} & {HPLT} & \multirow{2}{*}{I} & \multirow{2}{*}{II} & \multirow{2}{*}{III} & {HPLT} & {HPLT} & {HPLT} \\[-0.5ex]
&  & & & 1.2 & 2.0 & 3.0 & & & & 1.2 & 2.0 & 3.0 & & & & 1.2 & 2.0 & 3.0 & & & & 1.2 & 2.0 & 3.0 \\

\midrule
als\_Latn & - & - & - & - & - & - & - & - & - & - & - & - & - & - & - & - & - & - & 92.3 & 92.9 & 93.1 & 92.4 & \textbf{93.9} & 91.7 \\
ara\_Arab & 94.3 & \textbf{95.2} & \textbf{95.2} & 95.1 & - & 94.7 & 94.5 & 94.7 & 95 & \textbf{95.2} & - & 94.2 & 84.7 & 85.7 & \textbf{87.7} & 86.1 & - & 87.2 & 86.9 & 87.7 & 87.4 & \textbf{88.2} & - & 88.1 \\
bel\_Cyrl & 94.1 & 94.6 & 94.5 & 95.5 & \textbf{95.7} & 95.5 & 93.2 & 93.8 & 93.5 & 93.8 & \textbf{97.1} & 93.9 & 88.1 & 89.9 & 90.2 & 91.1 & \textbf{91.7} & 91.5 & 91.7 & 90.3 & 89.5 & 90.1 & \textbf{92.8} & 90.9\\
bul\_Cyrl & 97.0 & 97.5 & 97.6 & 97.8 & \textbf{97.9} & 97.9 & 97.5 & 97.7 & \textbf{98.0} & 97.3 & 97.3 & 96.9 & 92.7 & 94.4 & 94.2 & 94.0 & 94.5 & \textbf{94.8} & 92.2 & 92.2 & 91.1 & 91.5 & \textbf{93.0} & 91.7\\
cat\_Latn & 97.1 & 97.2 & 97.2 & 97.4 & \textbf{97.5} & 97.5 & \textbf{99.4} & 99.4 & 99.4 & 99.4 & 97.5 & 99.3 & 93.6 & 94.1 & 94.2 & 94.4 & \textbf{99.4} & 94.5 & 92.1 & 91.0 & 91.3 & 90.1 & \textbf{94.5} & 90.7\\
ces\_Latn & 97.8 & 98.0 & 98.1 & 98.3 & 98.4 & \textbf{98.5} & 99.3 & 99.3 & 99.4 & 99.4 & 99.4 & \textbf{99.5} & 93.5 & 94.2 & 94.1 & 94.4 & 94.6 & \textbf{94.8} & 91.2 & 91.2 & 90.4 & 89.0 & \textbf{91.8} & \textbf{91.8} \\
cym\_Latn & 87.2 & 88.3 & 88.7 & \textbf{89.2} & 89.0 & 88.8 & \textbf{94.6} & 94.4 & 94.6 & 93.7 & 92.3 & 92.6 & 80.8 & 82.8 & 83.2 & 82.3 & 82.8 & \textbf{83.6} & 92.5 & 90.0 & 92.4 & 89.4 & \textbf{93.4} & 91.0\\
dan\_Latn & 96.7 & 97.8 & 97.6 & 97.8 & 97.9 & \textbf{98.0} & 97.2 & \textbf{97.6} & 97.6 & 97.1 & 97.1 & 96.6 & 86.7 & 89.1 & 89.3 & 88.8 & 89.5 & \textbf{89.6} & 91.2 & 91.6 & 91.2 & 90.3 & 92.0 & 90.9\\
deu\_Latn & 88.8 & 89.4 & 89.5 & 80.7 & \textbf{89.9} & 89.8 & 97.6 & \textbf{97.7} & 97.6 & 95.5 & 97.5 & 97.6 & 84.6 & 87.1 & 86.9 & 76.4 & \textbf{87.6} & 87.4 & \textbf{89.4} & 87.7 & 88.6 & 64.1 & 89.2 & 88.8\\
ell\_Grek & 94.6 & 95.7 & 96.2 & 96.1 & 96.2 & \textbf{96.5} & 94.6 & 94.7 & \textbf{95.3} & 94.1 & 94.1 & 94.1 & 91.7 & 93.5 & 93.5 & 92.2 & 93.2 & \textbf{93.6} & 90.2 & 90.7 & 88.9 & 90.2 & \textbf{92.6} & 91.3\\
eng\_Latn & 96.1 & 96.8 & 97.0 & 96.7 & 97.0 & \textbf{97.1} & 97.8 & 98.0 & 98.0 & 97.9 & \textbf{98.1} & 98.0 & 91.3 & 92.6 & 93.0 & 92.2 & 93.0 & \textbf{93.3} & 2.2 & 81.1 & 82.3 & 81.0 & \textbf{82.7} & 80.1\\
spa\_Latn & 95.7 & 95.9 & 96.1 & 96.0 & \textbf{96.2} & 96.1 & \textbf{99.4} & 99.4 & 99.4 & 99.4 & 99.4 & 99.4 & 92.3 & 93.0 & 93.1 & 93.1 & \textbf{93.4} & 93.2 & \textbf{90.9} & 89.9 & 89.9 & 89.6 & 90.8 & 89.7\\
ekk\_Latn & 96.0 & 96.6 & 96.6 & \textbf{97.1} & 97.1 & 97.1 & 94.8 & 95.0 & \textbf{95.7} & 95.2 & 95.2 & 95.0 & 88.1 & 89.7 & 90.2 & 90.8 & \textbf{91.0} & 91.0 & 91.8 & 90.4 & 91.3 & 89.6 & \textbf{93.0} & 92.2\\
eus\_Latn & 91.0 & 91.4 & 92.1 & 92.3 & 92.3 & \textbf{92.4} & 95.7 & 95.9 & \textbf{96.3} & 96.0 & 95.9 & 95.9 & 85.3 & 87.3 & 87.3 & 88.1 & 88.2 & \textbf{88.5} & 91.3 & 90.7 & 89.6 & 89.8 & \textbf{92.9} & 92.4\\
fin\_Latn & 95.1 & 96.4 & 96.3 & 96.8 & \textbf{97.0} & 96.9 & 90.6 & 91.5 & \textbf{92.5} & 91.6 & 91.4 & 91.2 & 90.2 & 93.0 & 93.1 & 93.3 & \textbf{94.0} & 94.0 & 90.2 & 90.0 & 91.0 & 89.2 & \textbf{91.6} & 90.2\\
fra\_Latn & 97.8 & \textbf{98.1} & 98.0 & 98.1 & 98.0 & 98.1 & 98.6 & \textbf{98.8} & 98.8 & 93.8 & 98.6 & 98.5 & 93.8 & 94.4 & 94.7 & 94.5 & \textbf{94.8} & 94.8 & \textbf{90.5} & 88.7 & 89.4 & 87.2 & 90.0 & 88.6\\
gle\_Latn & 86.5 & 87.1 & 87.9 & 88.7 & 89.3 & \textbf{89.7} & 95.5 & 95.8 & 95.7 & \textbf{96.1} & 95.6 & 95.7 & 81.3 & 82.7 & 83.6 & 83.4 & \textbf{84.3} & 84.3 & \textbf{80.8} & 78.0 & 74.4 & 55.9 & 78.2 & 81.6\\
glg\_Latn & 96.9 & \textbf{97.1} & 97.0 & 97.1 & 97.0 & 96.8 & 98.3 & 98.3 & \textbf{98.4} & 98.2 & 98.0 & 97.9 & 82.3 & \textbf{82.6} & 82.1 & 82.3 & 82.2 & 82.0 & 92.5 & 93.3 & 92.0 & 91.1 & \textbf{94.1} & 91.4\\
heb\_Hebr & 95.6 & 96.1 & 96.0 & 96.5 & \textbf{96.7} & 96.7 & 97.0 & \textbf{97.2} & 97.2 & 97.1 & 97.2 & 97.2 & 89.8 & 91.6 & 90.9 & 91.0 & 91.9 & \textbf{92.0} & 2.6 & 84.2 & 84.0 & 88.4 & \textbf{89.3} & 87.8\\
hun\_Latn & 93.0 & \textbf{94.3} & 94.1 & 93.0 & 94.1 & 94.0 & 93.0 & \textbf{94.3} & 94.1 & 93.0 & 92.3 & 91.1 & 84.3 & 86.7 & 86.6 & 82.4 & 86.1 & \textbf{87.3} & 92.2 & 91.9 & 91.5 & 92.8 & \textbf{93.1} & 91.6\\
hye\_Armn & 88.7 & 91.2 & 90.1 & 92.7 & 92.7 & \textbf{93.4} & 94.4 & 94.9 & \textbf{95.0} & 93.9 & 94.7 & 94.3 & 80.4 & 85.3 & 84.6 & 84.1 & 86.8 & \textbf{87.7} & 95.7 & 95.3 & 94.1 & 94.8 & \textbf{95.9} & 96.3\\
ind\_Latn & 89.5 & 89.8 & \textbf{89.9} & 89.6 & 89.1 & 89.3 & 98.2 & \textbf{98.3} & 98.3 & 98.0 & 97.5 & 97.8 & 82.4 & \textbf{82.7} & 81.9 & 81.7 & 81.8 & 82.7 & 91.3 & 91.6 & 91.4 & 89.1 & \textbf{92.0} & 89.0\\
isl\_Latn & 87.7 & 88.1 & 88.3 & 88.6 & \textbf{88.7} & 88.5 & 96.2 & 96.4 & 96.4 & \textbf{96.5} & 96.4 & 95.8 & 85.2 & 86.6 & 86.7 & 86.9 & \textbf{87.4} & 86.4 & \textbf{81.7} & 63.9 & 81.3 & 55.9 & 78.3 & 81.7\\
ita\_Latn & 98.0 & 98.0 & 98.1 & 98.1 & \textbf{98.3} & 98.2 & 98.6 & 98.7 & 98.7 & \textbf{98.8} & 98.7 & 98.6 & 94.1 & 94.4 & 94.4 & 94.6 & \textbf{95.1} & 95.0 & 90.5 & 89.7 & 90.6 & 87.8 & \textbf{91.2} & 89.2\\
jpn\_Jpan & 97.5 & 97.7 & 97.7 & \textbf{97.8} & 97.8 & 97.6 & 98.3 & 98.3 & \textbf{98.4} & 98.3 & 98.4 & 98.2 & 94.1 & 94.6 & 94.7 & 94.6 & \textbf{94.8} & 94.8 & 66.5 & 65.9 & 65.7 & \textbf{67.4} & 67.2 & 67.1\\
kat\_Geor & 91.3 & \textbf{92.6} & 91.7 & 92.4 & 92.4 & 92.4 & 92.8 & 93.7 & \textbf{94.5} & 92.5 & 92.5 & 92.5 & 79.5 & 80.9 & 80.9 & 80.8 & 81.3 & \textbf{81.4} & 87.2 & 4.7 & 85.4 & 89.6 & \textbf{90.7} & 89.1\\
kor\_Hang & 88.6 & 89.7 & 89.5 & 89.9 & \textbf{90.1} & 89.9 & 94.0 & 94.3 & \textbf{94.4} & 94.4 & 94.4 & 94.3 & 88.0 & 89.0 & 89.1 & 89.4 & 89.7 & \textbf{89.9} & 87.8 & 87.0 & 86.3 & 88.3 & \textbf{89.3} & 87.6\\
lvs\_Latn & 91.6 & 92.8 & 92.7 & 92.4 & 93.6 & \textbf{93.7} & 96.9 & 91.6 & 97.5 & 96.8 & 97.7 & \textbf{97.8} & 88.8 & 90.9 & 91.1 & 90.9 & 92.1 & \textbf{92.4} & 93.2 & 92.6 & 91.2 & 90.7 & \textbf{93.9} & 92.4\\
lit\_Latn & 87.7 & 91.9 & 92.1 & 92.0 & 92.5 & \textbf{92.6} & 90.2 & 91.6 & \textbf{92.9} & 91.5 & 91.2 & 90.6 & 79.3 & 85.7 & 86.3 & 84.9 & \textbf{86.8} & 86.8 & 89.1 & 89.3 & 87.2 & 87.0 & \textbf{91.0} & 88.7\\
ltz\_Latn & - & - & - & - & - & - & - & - & - & - & - & - & - & - & - & - & - & - & 88.4 & 85 & 86.5 & - & \textbf{89.2} & 88.5 \\ 
mlt\_Latn & 94.7 & 94.5 & 97.1 & 97.0 & 97.7 & \textbf{97.8} & \textbf{100.0} & 100.0 & 100.0 & 100.0 & 100.0 & 100.0 & 78.2 & 78.5 & 86.6 & 83.2 & 87.2 & \textbf{88.5} & - & - & - & - & - & - \\
nob\_Latn & 97.0 & 97.4 & 97.4 & 97.6 & 97.5 & \textbf{97.7} & 98.5 & 98.8 & \textbf{98.9} & 98.8 & 98.7 & 98.8 & 93.2 & 94.3 & 94.1 & 94.5 & 94.7 & \textbf{94.8} & 91.9 & 92.6 & 91.7 & 91.1 & \textbf{93.2} & 92.1 \\
nld\_Latn & 96.2 & 96.9 & 96.7 & 97.1 & \textbf{97.2} & 96.9 & 94.1 & 94.7 & \textbf{94.9} & 94.4 & 94.1 & 94.1 & 91.6 & 92.9 & 93.6 & 93.8 & \textbf{94.1} & 94.0 & \textbf{91.7} & 90.4 & 90.5 & 88.6 & 91.0 & 90.0\\
nno\_Latn & 96.6 & 97.0 & 97.2 & 97.7 & \textbf{97.8} & 97.6 & 98.2 & 98.4 & \textbf{98.6} & 98.5 & 98.5 & 98.5 & 92.9 & 93.9 & 94.3 & 94.6 & \textbf{95.0} & 94.7 & \textbf{95.8} & 93.6 & 92.6 & 93.2 & 95.5 & 93.8\\
pol\_Latn & 95.6 & 95.5 & 96.6 & 96.9 & 97.2 & \textbf{97.3} & 97.8 & 98.2 & \textbf{98.3} & 98.2 & 98.2 & 98.2 & 93.7 & 95.2 & 95.3 & 95.3 & 95.6 & \textbf{95.7} & 12.9 & 88.8 & 89.0 & \textbf{89.7} & 89.6 & 89.1\\
por\_Latn & 93.6 & 94.0 & 93.9 & \textbf{94.1} & 94.1 & 94.1 & 98.1 & \textbf{98.3} & 98.2 & 98.3 & 98.2 & 98.2 & 83.4 & 84.5 & 85.0 & 84.9 & 85.3 & \textbf{85.4} & 91.2 & 90.3 & 88.9 & 88.0 & \textbf{91.5} & 89.5\\
ron\_Latn & 97.3 & 97.6 & 97.6 & 97.7 & \textbf{97.9} & 97.7 & 97.7 & 97.9 & \textbf{98.1} & 97.8 & 97.8 & 97.9 & 89.5 & 91.0 & 91.2 & 90.6 & \textbf{91.6} & 91.6 & \textbf{94.5} & 93.6 & 92.6 & 91.2 & 93.6 & 92.7\\
rus\_Cyrl & 93.8 & 94.4 & 94.8 & 94.5 & 94.7 & \textbf{95.1} & 98.3 & 98.5 & 98.4 & \textbf{98.6} & 98.6 & 98.4 & 92.6 & 93.4 & 93.6 & 93.6 & 93.8 & \textbf{94.0} & 88.0 & 86.9 & 85.5 & 85.6 & \textbf{89.0} & 87.4\\
slk\_Latn & 89.1 & 97.6 & 90.9 & \textbf{98.1} & 91.9 & 91.6 & 95.7 & 96.1 & \textbf{96.6} & 95.6 & 95.5 & 95.3 & 92.9 & 94.4 & 94.6 & 93.8 & \textbf{95.0} & 94.8 & 93.2 & 92.9 & 93.1 & 91.2 & \textbf{93.3} & 92.6\\
slv\_Latn & 96.7 & 97.6 & 97.6 & 98.1 & 98.2 & \textbf{98.3} & 98.5 & \textbf{98.7} & 98.7 & 98.6 & 98.7 & 98.3 & 93.4 & 94.7 & 94.8 & 94.8 & 95.3 & \textbf{95.4} & 93.4 & 93.1 & 92.5 & 93.6 & \textbf{94.2} & 92.7\\
srp\_Cyrl & - & - & - & - & - & - & - & - & - & - & - & - & - & - & - & - & - & - & 91.6 & 92.4 & 90.7 & - & \textbf{93.4} & 91.4 \\
swe\_Latn & 96.5 & \textbf{97.4} & 97.3 & 97.4 & 97.3 & 97.4 & 97.3 & 97.6 & \textbf{97.7} & 97.1 & 97.0 & 96.8 & 89.4 & 92.1 & 92.1 & 90.8 & 91.7 & \textbf{92.3} & 94.3 & \textbf{94.5} & \textbf{94.5} & 93.5 & 94.4 & 93.9\\
tam\_Taml & 79.6 & 80.9 & 80.9 & \textbf{82.8} & - & 80.2 & 87.9 & 89.7 & \textbf{89.3} & 88.6 & - & 80.3 & 62.9 & 64.9 & 61.1 & 63.6 & - & \textbf{65} & 84.2 & 84.5 & 85.4 & - & - & \textbf{88.4} \\
tat\_Cyrl & - & - & - & - & - & - & - & - & - & - & - & - & - & - & - & - & - & - & \textbf{89.7} & 80.6 & 84.1 & 82.9 & 84 & 86.4 \\
tha\_Thai & 90.2 & 91.6 & 91.4 & - & - & \textbf{91.7} & \textbf{100} & \textbf{100} & \textbf{100} & - & - & \textbf{100} & 73.8 & \textbf{78} & 77.8 - & - & - & 77.8 & 69.3 & \textbf{70.7} & 69.2 & - & - & 69 \\
tur\_Latn & 90.4 & 91.0 & 91.2 & \textbf{91.5} & 91.4 & 91.1 & 91.1 & 91.3 & 91.6 & \textbf{91.9} & 91.4 & 91.5 & 70.9 & 73.0 & 73.6 & 73.6 & 74.6 & \textbf{74.9} & 92.2 & 92.0 & 92.3 & 90.8 & \textbf{92.5} & 91.1\\
ukr\_Cyrl & 93.1 & 94.7 & 94.9 & 72.9 & \textbf{95.3} & 95.2 & 87.0 & 97.2 & \textbf{97.5} & 87.0 & 97.0 & 96.5 & 89.4 & 91.8 & 91.6 & 61.3 & 92.1 & \textbf{92.3} & 92.0 & 91.7 & 91.3 & 77.5 & \textbf{92.8} & 91.3\\
vie\_Latn & 89.8 & \textbf{92.1} & 91.5 & 91.8 & 92.1 & 92.1 & \textbf{99.9} & 99.9 & 99.9 & 99.9 & 99.9 & 99.9 & 66.5 & 70.3 & \textbf{71.3} & 68.0 & 70.3 & 71.0 & \textbf{91.9} & 90.6 & 89.8 & 89.2 & 90.3 & 89.5\\
cmn\_Hans & 96.2 & \textbf{96.3} & 96.3 & 96.0 & 96.0 & 95.8 & \textbf{99.9} & 99.9 & 99.9 & 99.9 & 99.9 & 99.9 & 86.1 & \textbf{86.9} & 86.9 & 84.6 & 85.6 & 85.8 & 0.1 & \textbf{76.5} & 76.5 & 75.5 & 74.5 & 75.6\\
\bottomrule
\end{tabular}
}
\caption{Results of monolingual GPT-BERT models compared to the baselines -- where ``I'' denotes mBERT, ``II'' XLM-R, and ``III'' mmBERT -- on part-of-speech (POS) tagging, lemmatization, dependency parsing and named entity recognition. For POS tagging, we evaluate the AllTags performance, which is the exact
match accuracy of the UPOS, XPOS, and UFeats UDtags. For dependency parsing, we report LAS, and for lemmatization accuracy.}
\label{tab:mlm}
\end{table*}

\begin{table*}[!htp]
\centering
\small
\adjustbox{max width=\linewidth}{
\begin{tabular}{ll|ccccc|cc}
\toprule
\textbf{} & \textbf{} & \multicolumn{5}{c|}{\textbf{MultiBLIMP BERT}}  & \multicolumn{2}{c}{\textbf{MultiBLIMP GPT}} \\
\textbf{ISO-639} & \textbf{Language} & XLM-R & mmBERT & HPLT 1.2 & HPLT 2.0 & \HPLT{} & Goldfish & \HPLT{} \\
\midrule
\textbf{als\_Latn} & Albanian & 93.8 & 92.2 & 93.8 & 93.8 & \textbf{97.9} & \textbf{99.2} & 97.1 \\
\textbf{ara\_Arab} & Arabic & 92 & 93.3 & 77.2 & - & \textbf{97.4} & 95.2 & \textbf{95.9} \\
\textbf{bel\_Cyrl} & Belarusian & 94.4 & 91.9 & 81.8 & 82.2 & \textbf{99.3} & 97.3 & \textbf{98.3} \\
\textbf{bul\_Cyrl} & Bulgarian & 96.5 & 96.5 & 85.8 & 85.6 & \textbf{99.5} & 97.2 & \textbf{98.5} \\
\textbf{cat\_Latn} & Catalan & 94.7 & 94.4 & 88.3 & 88.6 & \textbf{98.4} & 97.7 & \textbf{97.9} \\
\textbf{ces\_Latn} & Czech & 96.5 & 97.2 & 87 & 87.2 & \textbf{99.1} & 92.2 & \textbf{96.7} \\
\textbf{cym\_Latn} & Welsh & 87.6 & 86.8 & 93.1 & 93.5 & \textbf{99.6} & 92.2 & \textbf{97.9} \\
\textbf{dan\_Latn} & Danish & 98 & 96 & 84 & 86 & \textbf{100} & \textbf{100} & \textbf{100} \\
\textbf{deu\_Latn} & German & 97.7 & 98.9 & 55.4 & 90.7 & \textbf{99.1} & 98 & \textbf{98.3} \\
\textbf{eng\_Latn} & English & 96.9 & \textbf{97.3} & 88.6 & 90.3 & 96.1 & 96.4 & \textbf{97.7} \\
\textbf{ekk\_Latn} & Estonian & 95.3 & 94.8 & 86.5 & 86.6 & \textbf{99.5} & 96.2 & \textbf{97.4} \\
\textbf{ell\_Grek} & Greek & 97.8 & 96.9 & 94.3 & 94.3 & \textbf{99.6} & 98.8 & \textbf{99.1} \\
\textbf{eus\_Latn} & Basque & 94.9 & 93 & 96.7 & 97.8 & \textbf{99.3} & \textbf{98.9} & 98.2 \\
\textbf{fao\_Latn} & Faroese & 72.8 & 87.9 & - & \textbf{99.1} & 98.3 & \textbf{99.6} & 98.3 \\
\textbf{fin\_Latn} & Finnish & 96.3 & 93.6 & 97.4 & 97.7 & \textbf{99.3} & 96.2 & \textbf{97.7} \\
\textbf{fra\_Latn} & French & 97 & 97.4 & 93 & 92.9 & \textbf{98.7} & 98.5 & \textbf{99.5} \\
\textbf{gle\_Latn} & Irish & 71.4 & 71.4 & 71.4 & \textbf{85.7} & \textbf{85.7} & \textbf{92.9} & 82.1 \\
\textbf{glg\_Latn} & Galician & 93.1 & 94.7 & 89.6 & 89.5 & \textbf{98.5} & \textbf{98.3} & 97.1 \\
\textbf{heb\_Hebr} & Hebrew & 90.3 & 91.9 & 71.4 & 74.5 & \textbf{95.9} & 85.9 & \textbf{87.3} \\
\textbf{hun\_Latn} & Hungarian & 98.3 & 98.5 & 87.9 & 87.9 & \textbf{99.8} & 97.5 & \textbf{98.9} \\
\textbf{hye\_Armn} & Armenian & 98.5 & - & 99.4 & 99.4 & \textbf{99.6} & 98.4 & \textbf{98.5} \\
\textbf{isl\_Latn} & Icelandic & 91.5 & 93 & 98.9 & 99.2 & \textbf{99.5} & \textbf{98.4} & 98.3 \\
\textbf{ita\_Latn} & Italian & 93.4 & 94.1 & 85.8 & 86.3 & \textbf{97.9} & 94.6 & \textbf{96.7} \\
\textbf{kat\_Geor} & Georgian & 89.7 & - & 94.1 & 94.6 & \textbf{95.1} & 96.6 & \textbf{97.5} \\
\textbf{kmr\_Latn} & Northern Kurdish & 80.9 & 84.2 & - & - & \textbf{98.7} & 94.7 & \textbf{97.2} \\
\textbf{lit\_Latn} & Lithuanian & 95.3 & 94.3 & 97.8 & 98 & \textbf{99.7} & 97.9 & \textbf{98.7} \\
\textbf{lvs\_Latn} & Latvian & 93.6 & 92.6 & 64.1 & 92.2 & \textbf{99.5} & 96.8 & \textbf{98} \\
\textbf{mkd\_Cyrl} & Macedonian & 94.9 & 94.9 & 46.2 & \textbf{100} & \textbf{100} & \textbf{100} &  \textbf{100} \\
\textbf{nob\_Latn} & Norwegian Bokmål & 89.7 & 87.7 & 88.9 & 89.6 & \textbf{93.4} & \textbf{91} & \textbf{91} \\
\textbf{nld\_Latn} & Dutch & 95.9 & 97.3 & 93 & 92.4 & \textbf{98.9} & 97.3 & \textbf{98.4} \\
\textbf{pol\_Latn} & Polish & 96.7 & 96.9 & 91.7 & 91.9 & \textbf{99.4} & 96.3 & \textbf{97.6} \\
\textbf{por\_Latn} & Portuguese & 96.1 & 96.2 & 94.2 & 94 & \textbf{98.2} & 94.4 & \textbf{95.7} \\
\textbf{ron\_Latn} & Romanian & 96.3 & 96 & 93.6 & 94.2 & \textbf{99.2} & 96.5 & \textbf{97.8} \\
\textbf{rus\_Cyrl} & Russian & 97.3 & 98.3 & 79.4 & 79.4 & \textbf{99.2} & 94.5 & \textbf{97.1} \\
\textbf{slk\_Latn} & Slovak & 94.5 & 95.4 & 95.7 & 95.7 & \textbf{99.6} & 95.2 & \textbf{97.1} \\
\textbf{slv\_Latn} & Slovene & 94.2 & 93.7 & 94.2 & 94.2 & \textbf{98.7} & 93.6 & \textbf{95.9} \\
\textbf{spa\_Latn} & Spanish & 96.3 & 96.4 & 75.1 & 75.7 & \textbf{98} & 96.1 & \textbf{97} \\
\textbf{swe\_Latn}  & Swedish & 99.5 & \textbf{100} & 94.5 & 94.5 & 99 & \textbf{100} & \textbf{100} \\
\textbf{tam\_Taml} & Tamil & 94.8 & 96.6 & \textbf{99.5} & - & \textbf{99.5} & 98.2 & \textbf{98.4} \\
\textbf{tur\_Latn} & Turkish & 91 & 90.5 &  88.3 & 89.3 & \textbf{98} & 93.6 & \textbf{96.9} \\
\textbf{ukr\_Cyrl} & Ukrainian & 97 & 97 & 53.4 & 89.6 & \textbf{99.1} & 95.9 & \textbf{97.4} \\
\bottomrule
\end{tabular}
}
\caption{Evaluation results of \HPLT{} monolingual GPT-BERTs on MultiBLIMP (NoCOLA). We do not report mmBERT's performance for Armenian and Georgian because high fertility of its tokenizer for these languages caused out of memory issues. `MultiBLIMP BERT' denotes inference as a masked language model, and `MultiBLIMP GPT' as a causal one.}
\label{tab:mlm-multiblimp}
\end{table*}
\begin{table*}[t!]
  \centering\smaller
  \tabcolsep 0.3em
  \begin{tabular*}{\textwidth}{@{\extracolsep{\fill}}lrccrccc}
    \toprule
    \textbf{Language} 
    & \multicolumn{3}{c}{\bf Named Entity Recognition \textbf{(WikiAnn, F1)}}
    & \multicolumn{4}{c}{\bf Linguistic Competence (\textbf{MultiBLIMP, Acc)}}\\[0.5ex]
     & \textbf{Size} & \textbf{\HPLT} &
     \textbf{mT5-base} &
    \textbf{Size} & \textbf{\HPLT} 
    & \textbf{mT5-base} & \textbf{mT5-xxl} \\
    \cmidrule(r){1-1}\cmidrule(rl){2-4}\cmidrule(l){5-8}
    Albanian (\textbf{als\_Latn}) & 100 & 93.2 & 86.7 & 243 & 95.5 & 90.5 & 88.9 \\
    Arabic (\textbf{ara\_Arab})  & 10000 & 91.7 & 80.8 & 1215 & 92.4 & 87.7 & 95.1 \\
    Asturian (\textbf{ast\_Latn}) & 1000 & 89.4 & 60.2 & - & - & - & - \\
    Belarusian (\textbf{bel\_Cyrl}) & 1000 & 91.5 & 86 & 2570 & 97.2 & 84.5 & 90.3 \\
    Bosnian (\textbf{bos\_Latn}) & 1000 & 94.2 & 88.4 & 3286 & 92.2 & 78.6 & 92 \\
    Bulgarian (\textbf{bul\_Cyrl}) & 10000 & 93.3 & 78.6 & 2458 & 93 & 87.7 & 91.6 \\
    Catalan (\textbf{cat\_Latn}) & 10000 & 92.7 & 87.4 & 2284 & 95.6 & 91.6 & 93.0 \\
    Czech (\textbf{ces\_Latn}) & 10000 & 91.6 & 85.2 & 4256 & 95.9 & 88.8 & 93.4 \\
    Chinese (\textbf{cmn\_Hans}) & 10000 & 80.5 & 70.6 & - & - & - & - \\
    Welsh (\textbf{cym\_Latn}) & 1000 & 93.6 & 81.4 & 1120 & 89.3 & 78.1 & 86.1 \\
    Danish (\textbf{dan\_Latn}) & 10000 & 91.6 & 87.5 & 50 & 100 & 98 & 96 \\
    German (\textbf{deu\_Latn}) & 10000 & 88.6 & 83.4 & 2298 & 96 & 94 & 97 \\
    English (\textbf{eng\_Latn}) & 10000 & 82.1 & 77.6 & 770 & 94.2 & 90.6 & 95.3 \\
    Estonian (\textbf{ekk\_Latn})  & 10000 & 92 & 81.1 & 2575 & 97.3 & 82.6 & 85.7 \\
    Greek (\textbf{ell\_Grek}) & 10000 & 92.5 & 86.1 & 1096 &  98.5 & 96.4 & 98.3 \\
    Basque (\textbf{eus\_Latn}) & 10000 & 92.0 & 82.8 & 273 & 97.4 & 94.9 & 96.0 \\
    Faroese (\textbf{fao\_Latn}) & 100 & - & - & 232 & 95.7 & 71.6 & 85.3 \\
    Finnish (\textbf{fin\_Latn}) & 10000 & 90.3 & \phantom{0}1.8 & 2570 & 95.6 & 81.4 & 86.1 \\
    French (\textbf{fra\_Latn}) & 10000 & 88.9 & 83.3 & 2548 & 93.6 & 91.7 & 94.8 \\
    Japanese (\textbf{jpn\_Jpan}) & 10000 & 73.6 & 54.3 & - & - & - & - \\
    Irish (\textbf{gle\_Latn}) & 1000 & 82.1 & 60.1 & 28 & 89.3 & 53.6 & 78.6 \\
    Galician (\textbf{glg\_Latn}) & 10000 & 93.4 & 89.2 & 753 & 96.0 & 90.7 & 95.4 \\
    Hebrew (\textbf{heb\_Hebr}) & 10000 & 88.9 & 77.1 & 2330 & 82.4 & 79.6 & 90.6 \\
    Croatian (\textbf{hrv\_Latn}) & 10000 & 91.4 & 86.8 & 3286 & 92.8 & 78.6 & 92 \\
    Hungarian (\textbf{hun\_Latn}) & 10000 & 91.9 & 84.9 & 845 & 99.1 & 92.8 & 95.9 \\
    Armenian (\textbf{hye\_Armn}) & 1000 & 96.2 & 89.5 & 1415 & 90.2 & 89.5 & 92.2 \\
    Indonesian (\textbf{ind\_Latn}) & 10000 & 92.4 & 85.9 & - & - & - & - \\
    Icelandic (\textbf{isl\_Latn}) & 1000 & 83.8 & 71 & 2801 & 94 & 87.3 & 91.1 \\
    Italian (\textbf{ita\_Latn}) & 10000 & 90.9 & 85.4 & 2999 & 93.9 & 88.5 & 94.7 \\
    Georgian (\textbf{kat\_Geor}) & 10000 & 90.4 & 80.4 & 204 & 96.6 & 93.6 & 90.7 \\
    Korean (\textbf{kor\_Hang}) & 10000 & 85.9 & 79.5 & - & - & - & - \\
    Northern Kurdish (\textbf{kmr\_Latn}) & 100 & - & - & 544 & 94.7 & 77 & 84 \\
    Lithuanian (\textbf{lit\_Latn}) & 10000 & 90 & 84.5 & 1180 & 98 & 92.2& 87.7 \\
    Luxembourgish (\textbf{ltz\_Latn}) & 1000 & 88.6 & 4 & - & - & - & - \\
    Latvian (\textbf{lvs\_Latn}) & 10000 & 92.9 & 86.1 & 3032 & 96.4 & 84 & 87.3 \\
    Macedonian (\textbf{mkd\_Cyrl}) & 1000 & 93.8 & 78.3 & 39 & 100 & 94.9 & 92.3 \\
    Dutch (\textbf{nld\_Latn}) & 10000 & 90.7 & 85.6 & 2331 & 92.1 & 89.3 & 94.1\\
    Bokmål (\textbf{nob\_Latn}) & 10000 & 91.8 & 87.0 & \textbf{*}3463 & 40.6 & 68.0 & 71.8 \\
    Nynorsk (\textbf{nno\_Latn})  & 1000 & 94.0 & 88.2 & - & - & - & - \\
    Polish (\textbf{pol\_Latn}) & 10000 & 89.6 & 87.8 & 3272 & 94.9 & 86.6 & 89.3 \\
    Portuguese (\textbf{por\_Latn}) & 10000 & 91.3 & 89.9 & 3048 & 93.5 & 92 & 95 \\
    Romanian (\textbf{ron\_Latn}) & 10000 & 93.6 & 86.4 & 2056 & 91.3 & 86.9 & 91.8 \\
    Russian (\textbf{rus\_Cyrl}) & 10000 & 88.2 & 82.9 & 3832 & 96.3 & 93 & 96.7 \\
    Slovak (\textbf{slk\_Latn}) & 10000 & 92.9 & 88.8 & 4145 & 92.8 & 80.2 & 86.6 \\
    Slovene (\textbf{slv\_Latn}) & 10000 & 92.5 & 86.4 & 4483 & 92.6 & 83.6 & 90 \\
    Spanish (\textbf{spa\_Latn})  & 10000 & 90.7 & 84.0 & 2541 & 95.2 & 93.8 & 96.3 \\
    Serbian (\textbf{srp\_Cyrl}) & 10000 & 92.6 & 83.5 & - & - & - & - \\
    Swedish (\textbf{swe\_Latn})  & 10000 & 94.5 & 91.5 & 201 & 99.5 & 100 & 100 \\
    Swahili (\textbf{swh\_Latn}) & 1000 & 89.2 & 79.8 & - & - & - & - \\
    Tamil (\textbf{tam\_Taml}) & 1000 & 90 & 81 & 382 & 98.7 & 95.5 & 96.3 \\
    Thai (\textbf{tha\_Thai}) & 10000 & 80 & 32.1 & - & - & - & - \\
    Turkish (\textbf{tur\_Latn}) & 10000 & 92.3 & 87.9 & 1742 & 96.4 & 85.2 & 89.7 \\
    Ukrainian (\textbf{ukr\_Cyrl}) & 10000 & 92.5 & 82.1 & 2744 & 95.7 & 89.4 & 94.8 \\
    Vietnamese (\textbf{vie\_Latn}) & 10000 & 91.5 & 58.1 & - & - & - & - \\
    \cmidrule(r){1-1}\cmidrule(rl){2-4}\cmidrule(l){5-8}
    \textbf{Average} & - & 90.5 & 78.8 & - & 93.5 & 86.8 & 91.4 \\
    \bottomrule
  \end{tabular*}
  \caption{Evaluation results of \HPLT{} monolingual encoder--decoders (Bokmål and Nynorsk are two varieties of Norwegian), along with the test set sizes for each language. \textbf{*}Bokmål competence benchmarks are not part of MultiBLIMP.}
  \label{tb:t5_full}
\end{table*}

\section{Descriptive Statistics}
\label{ax:stats}

\begin{figure*}
    \centering
    \includegraphics[width=0.8\linewidth]{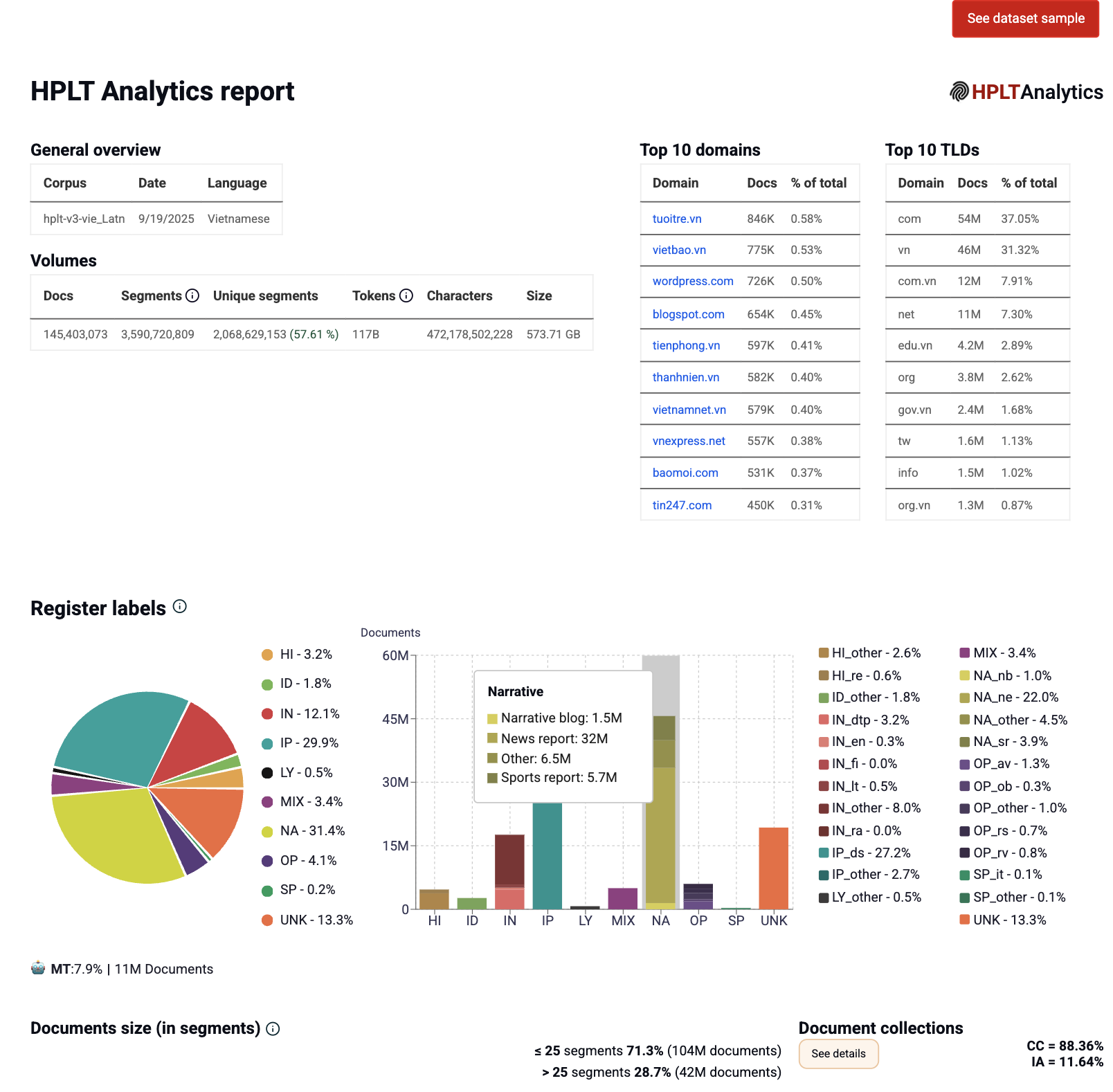}
    \caption{HPLT Analytics Dashboard screenshot showing statistics of the Vietnamese HPLT 3.0 dataset.}
    \label{fg:analytics-dashboard}
\end{figure*}

As mentioned in section~\ref{sc:analytics}, we run HPLT Analytics on HPLT 3.0 datasets and get in-depth insights from them. The tool creates interactive dashboards for each language as show in ~\ref{fg:analytics-dashboard}. 

Table \ref{tab:mono_unique} exhibits the languages with the highest and lowest increases in unique segments after comparing HPLT 2.0 and \HPLT{}, as well as the languages with the lowest ratios of unique segments.

Table \ref{tab:mono_large} shows languages with the highest and lowest ratios of large documents (> 25 segments)

Finally, table \ref{tab:mono_langid} shows the datasets with higher ratios of segments in the document language in two categories (large datasets and languages with infrequent writing systems) and languages that are confused with other larger, higher-resourced ones.



\begin{table}[tp]
    \adjustbox{max width=\linewidth}{
    \footnotesize
    \begin{tabular}{lrr}
    \toprule
    \multirow{2}{*}{\textbf{Language}} & \textbf{\% of Unique} & \textbf{\% of Unique}\\
    & \textbf{Segments (HPLT 2.0)} & \textbf{Segments (\HPLT{})}  \\
    \midrule
     \rowcolor{gray!20} 
     \multicolumn{3}{c}{Highest Increases} \\
     \midrule
    Nuer & 32\% & 98\% \\ 
    Wolof & 36\% & 89\% \\ 
    Dzonghka & 46\% & 93\% \\ 
    Ewe & 43\% & 90\% \\ 
    Dinka & 45\% & 90\% \\ 
    Waray & 43\% & 87\% \\ 
    Iloko & 33\% & 76\% \\ 
    Esperanto & 35\% & 77\% \\ 
    Limburgish & 38\% & 79\& \\ 
    Occitan & 30\% & 70\% \\ 
  \midrule
    \rowcolor{gray!20} 
    \multicolumn{3}{c}{Lowest Increases} \\
  \midrule
    Swati & 63\% & 62\% \\ 
    Balinese & 27\% & 28\% \\ 
    North Azerbaijani & 40\% & 43\% \\ 
    Manipuri & 61\% & 64\% \\ 
    Southern Sotho & 74\% & 78\% \\ 
    Friulian & 37\% & 43\% \\ 
    Xhosa & 58\% & 64\% \\ 
    Bengali & 56\% & 63\% \\ 
    Tatar & 47\% & 54\% \\ 
    Khmer & 59\% & 67\% \\ 
    Shona & 72\% & 80\% \\ 
    \midrule
    \rowcolor{gray!20} 
    \multicolumn{3}{c}{Lowest Ratios in \HPLT{}} \\
  \midrule
    Balinese & 27\% & 28\% \\
    North Azerbaijani & 40\% & 43\% \\
    Friulian & 37\% & 43\% \\
    Czech & 34\% & 47\% \\
    Polish & 33\% & 47\% \\
    French & - & 48\% \\
    Estonian & 39\% & 50\% \\
    Hungarian & 35\% & 50\% \\
    
    \bottomrule
    \end{tabular}}
    \caption{Different values of unique segments ratios.}
    \label{tab:mono_unique}
\end{table}

\begin{table}[tp]
    \adjustbox{max width=\linewidth}{
    \footnotesize
    \begin{tabular}{lr}
    \toprule
    \multirow{2}{*}{\textbf{Language}} & \textbf{\% of Large} \\
    & \textbf{Segments}  \\
    \midrule
     \rowcolor{gray!20} 
     \multicolumn{2}{c}{Highest Ratios} \\
     \midrule
    Dzongkha & 90\% \\
    Tamasheq (Tifinagh) & 60\% \\ 
    Balinese & 49\% \\ 
    Dyula & 39\% \\
    Magahi & 39\% \\
    Korean & 36\% \\
    Luba-Lulua & 36\% \\
    Sanskrit & 36\% \\
    Chokwe & 35\% \\
    Swati & 35\% \\
    Akan & 35\% \\
  \midrule
    \rowcolor{gray!20} 
    \multicolumn{2}{c}{Lowest ratios} \\
  \midrule
    Kashmiri (Devanagari) & 2\% \\
    Odia & 4\% \\
    Central Kanuri (Latin) & 5\% \\
    Awadhi & 6\% \\
    Crimean Tatar & 6\% \\
    Sardinian & 7\% \\
    Lao & 8\% \\
    Lombard & 8\%  \\
    Malayalam & 8\% \\
 
    \bottomrule
    \end{tabular}}
    \caption{Different values of large segment ratios in HPLT 3.0.}
    \label{tab:mono_large}
\end{table}

\begin{table}[tp]
    \adjustbox{max width=\linewidth}{
    \footnotesize
    \begin{tabular}{llr}
    \toprule
    \textbf{Language} & \textbf{Identified Language} & \textbf{\% of Segments}\\  
    \midrule
     \rowcolor{gray!20} 
     \multicolumn{3}{c}{High-Resourced Languages} \\
     \midrule  
    Russian* & Russian & 94\% \\
    English* & English & 94\% \\
    Ukrainian & Ukrainian & 92\% \\
    Belarussian & Belarussian & 90\% \\
    Hungarian & Hungarian & 90\% \\
    Kazakh & Kazakh & 90\% \\
    Finnish & Finnish & 89\% \\
    Bulgarian & Bulgarian & 88\% \\
    Italian & Italian & 88\% \\
    French & French & 87\% \\
    Lithuanian & Lithuanian & 87\% \\
    Hindi & Hindi & 87\% \\
    Latvian & Latvian & 87\% \\
    Czech & Czech & 86\% \\    
    German & German & 86\% \\
    \midrule
     \rowcolor{gray!20} 
     \multicolumn{3}{c}{Infrequent Scripts} \\
     \midrule 
     Bengali  & Bengali & 95\% \\
     Hebrew & Hebrew & 95\% \\
     Georgian & Georgian & 93\% \\
     Khmer & Khmer & 93\% \\
     Thai & Thai & 93\% \\
     Lao & Lao & 92\% \\
     Gujarati & Gujarati & 91\% \\
     Greek & Greek & 92\% \\
     Kannada & Kannada & 91\%\\
     Amharic & Amharic & 91\% \\
     Armenian & Armenian &  90\% \\
     Japanese* & Japanese & 90\% \\
     Punjabi & Punjabi & 90\% \\
     Tamil & Tamil & 90\% \\
    \midrule
     \rowcolor{gray!20} 
     \multicolumn{3}{c}{Low-Resourced Languages} \\
     \midrule  
    Sicilian & Sicilian & 2\% \\
        & Italian & 76\% \\
    Yue & Yue & 2\% \\
        & Chinese & 76\% \\
    Minangkabau & Minangkabau & 3\% \\
        & Indonesian & 59\% \\
    Lombard & Lombard & 4\% \\
        & Italian & 30\% \\     
    Egyptian Arabic & Egyptian Arabic & 5\% \\ 
        & Arabic & 88\% \\
    Bosnian & Bosnian & 16\% \\
        & Serbian & 58\% \\
    Venetian & Venetian & 17\% \\
        & Italian & 58\% \\
    Awadhi & Awadhi & 24\% \\
        & Hindi & 65\% \\   
    Latgalian & Latgalian & 36\% \\
        & Latvian & 41\% \\

    \bottomrule
    \end{tabular}}
    \caption{Language identification at segment level. Statistics for Russian, English, and Japanese are based on a random sample.}
    \label{tab:mono_langid}
\end{table}

\section{Domains and Register Labels}
\label{ax:domains}


%
%

Table \ref{tab:mono_domains} shows examples of  patterns found regarding the domains from which the documents were crawled. We include the ratio of documents for HPLT 2.0 and \HPLT{}, that illustrates the  general decrease in documents from Wikipedia after our process of global deduplication.

\begin{table}[tp]
    \adjustbox{max width=\linewidth}{
    \footnotesize
    \begin{tabular}{lrr}
    \toprule
    \multirow{2}{*}{\textbf{Language}} & \textbf{\% of Documents}  & \textbf{\% of Documents}\\
    & \textbf{HPLT 2.0} & \textbf{\HPLT{}} \\
    \midrule
     \rowcolor{gray!20} 
     \multicolumn{3}{c}{Wikipedia} \\
     \midrule
    Santali & 90\% & 85\% \\
    Ligurian & 80\% & 76\% \\ 
    Kashmiri (Arabic) & 23\% & 46\% \\
    Esperanto & 66\% & 40\% \\
    Asturian & 54\% & 33\% \\
    Sanskrit & 42\% & 31\% \\
    Occitan & 62\% & 25\% \\
  \midrule
    \rowcolor{gray!20} 
    \multicolumn{3}{c}{Biblical Domains} \\
    \midrule
    Bemba & 72\% & 100\% \\
    Tamasheq (Latin) & 80\% & 99\% \\
    Chokwe & 92\% & 98\% \\
    Kamba & 92\% & 98\% \\
    Sango & 88\% & 93\% \\
    Dyula & 99\% & 90\% \\
    Tumbuka & 92\% & 90\% \\
    Kongo/Kituba & 75\%& 87\% \\
    \bottomrule
    \end{tabular}}
    \caption{Frequent domain types found in different versions and languages of HPLT monolingual datasets.}
    \label{tab:mono_domains}
\end{table}

Table \ref{tab:mono_rl} lists examples of the register labels distribution patterns  found in \HPLT{}, illustrating  observations from Section \ref{sc:analytics}, while Table \ref{tab:mono_rl_outliers} shows outliers of the patterns. Note that register labels are not available for all languages in \HPLT{}.

\begin{table}[tp]
    \adjustbox{max width=\linewidth}{
    \footnotesize
    \begin{tabular}{lr}
    \toprule
    \textbf{Language} & \textbf{\% of Documents} \\
    \midrule
     \rowcolor{gray!20} 
     \multicolumn{2}{c}{Narrative} \\
     \midrule
    Nepali & 84\% \\
    Odia & 81\%\\
    Malayalam & 79\% \\
    Somali & 77\% \\
    Bengali & 77\% \\
    Kannada & 73\% \\
  \midrule
    \rowcolor{gray!20} 
    \multicolumn{2}{c}{Informational Persuasion + Narrative} \\
    \midrule
    Czech & 37\% + 27\% \\ 
    Polish & 36\% + 28\%  \\
    Slovak & 36\% + 27\% \\
    Danish & 33\% + 24\% \\
    French & 32\% + 25\% \\
    Dutch & 32\% + 27\% \\
    German & 31\% + 27\% \\
 
    \bottomrule
    \end{tabular}}
    \caption{Frequent register labels patterns found in \HPLT{} datasets for different languages.}
    \label{tab:mono_rl}
\end{table}

\begin{table}[tp]
    \adjustbox{max width=\linewidth}{
    \footnotesize
    \begin{tabular}{llr}
    \toprule
    \multirow{2}{*}{\textbf{Language}}  &     \textbf{} & \textbf{\% of} \\
    & \textbf{Register (Subregister)} & \textbf{Documents} \\ 
    
    \midrule
     \rowcolor{gray!20} 
     \multicolumn{3}{c}{Outliers} \\
    \midrule
    English* & Informational Description (Thing or Person) &    33\% \\
        &  & \\
    Esperanto & Informational Description (Encyclopedia article) & 50\% \\
        &  & \\
    Sanskrit & Opinion (Religious blog / Sermon) & 34\% \\
        &  & \\
    Yiddish & Interactive Discussion (Other) & 20\% \\
        &  & \\
    \bottomrule
    \end{tabular}}
    \caption{Languages that do not follow the patterns frequently found in register labels distribution. Statistics for English are based on a random sample.}
    \label{tab:mono_rl_outliers}
\end{table}

\section{Geographic TLDs}
\label{ax:tlds}

Table \ref{tab:mono_tlds} lists examples of geographic TLDs in the \HPLT{} corpus that illustrate the observations included in Section \ref{sc:analytics}.
%

\begin{table}[tp]
    \adjustbox{max width=\linewidth}{
    \footnotesize
    \begin{tabular}{lrcl}
    \toprule
    \multirow{2}{*}{\textbf{Language}}  & \textbf{\% of} & \multirow{2}{*}{\textbf{TLD}}  & \textbf{Country or}\\
     & \textbf{Documents} & & \textbf{Territory}\\
    \midrule
     \rowcolor{gray!20} 
     \multicolumn{4}{c}{One Geographic TLD} \\
    \midrule
        Latgalian & 90\% & .lv & Latvia \\
        Faroese & 87\% & .fo & Faroe Islands \\
        Manipuri & 84\% & .in & India \\
        Lithuanian & 79\% & .lt & Lithuania \\
        Polish & 78\% & .pl & Poland \\
        Danish & 75\% & .dk & Denmark \\
        Luganda & 75\% & .ug  & Uganda \\
        Georgian & 74\% & .ge & Georgia \\
        Norwegian Nynorsk & 73\% & .no & Norway \\
        Macedonian & 72\% & .mk & North Macedonia \\
        Latvian & 71\% & .lv & Latvia \\

  \midrule
    \rowcolor{gray!20} 
    \multicolumn{4}{c}{Language Variants} \\
    \midrule
        Romanian & 74\% & .ro & Romania \\
                & 3\% & .md & Moldova \\
        Greek & 70\% & .gr & Greece \\
                & 1\% & .cy & Cyprus \\
        Dutch & 66\% & .nl & Netherlands \\
                & 12\% & .be & Belgium \\
        Lombard & 65\% & .ch & Switzerland \\
                & 10\% & .it & Italy \\                
        German & 57\% & .de & Germany \\
                & 6\% & .at & Austria \\
                & 5\% & .ch & Switzerland \\
        Italian & 57\% & .it & Italia \\
                & 1\% & .ch & Switzerland \\                
        Portuguese & 46\% & .br & Brazil \\
                & 8\% & .pt & Portugal \\
        French & 30\% & .fr & France \\
                & 3\% & .be & Belgium \\
                & 2\% & .ca & Canada \\
                & 2\% & .ch & Switzerland \\
        Rundi & 28\% & .rw & Rwanda \\
                & 8\% & .bi & Burundi \\
        Tswana & 23\% & .za & South Africa \\
                & 11\% & .bw & Botswana \\

    \midrule
    \rowcolor{gray!20} 
    \multicolumn{4}{c}{Related Territories} \\
    \midrule
        Czech & 83\% & .cz & Czechia \\
                & 1\% & .sk & Slovakia \\
        Slovak & 76\% & .sk & Slovakia \\
                & 3\% & .cz & Czechia \\
        Icelandic & 74\% & .is & Iceland \\
                & 2\% & .dk & Denmark \\
        Russian* & 64\% & .ru & Russia \\
                & 5\% & .ua  & Ukraine \\
                & 2\% & .by & Belarus \\
                & 1\% & .kz & Kazakhstan \\
        Moroccan Tamazight & 54\% & .ma & Morocco \\ 
                & 19\% & .dz & Algeria \\
        Ukrainian & 50\% & .ua & Ukraine \\
                & 2\% & .ru & Russia \\
        Tajik & 46\% & .uz & Uzbekistan \\
                & 15\% & .tj & Tajikistan \\
                & 1\% & .kz & Kazakhstan \\
                & 1\% & .ru & Russia \\
        Bosnian & 34\% & .rs & Serbia \\
                & 15\% & .ba & Bosnia \\ 
                & 4\% & .me & Montenegro \\
                & 2\% & .hr & Croatia \\

    \bottomrule
    \end{tabular}}
    \caption{Frequent geographic TLDs in monolingual \HPLT{} datasets for different languages. Statistics for Russian are based on a random sample.}
    \label{tab:mono_tlds}
\end{table}

\section{Frequent \textit{n}-grams}
\label{ax:ngrams}

Table \ref{tab:mono_ngrams} shows examples of patterns of \textit{n}-grams in \HPLT{} that might be indicators of low quality or biased text. Translations into English are obtained with Google Translate.\footnote{\url{translate.google.com}}

\begin{table*}[htbp]
    \centering
    \adjustbox{max width=\linewidth}{
    \begin{tabular}{lllr}
    \toprule
    \textbf{Language} & \textbf{\textit{N}-gram (Original)}  & \textbf{\textit{N}-gram (Translated)}   & \textbf{Occurrences}  \\
    \midrule
     \rowcolor{gray!20}      
     \multicolumn{4}{c}{Religious \& Biblical Content} \\
    \midrule
    Tok Pisin & god & - & 91K \\
    Ayacucho Quechua & diospa & of god & 54K \\
    Ewe & yehowa &  jehova & 42K \\
    Tumbuka & jehova & - & 33K \\
    Kituba & nzambi & god & 32K \\ 
    Bemba & yesu & jesus & 27K \\
    Tumbuka & chiuta & god & 26K \\
    Tumbuka & yesu & jesus & 22K \\
    Ewe & biblia & bible & 20K \\
    Bemba & yehova & jehova & 15K \\
    Chokwe & yehova & jehova & 10K \\
    Wolof & yeesu & jesus & 9K \\
    Luganda & mukama katonda & lord god & 7K \\
    Swati & nkulunkulu & god & 5K \\
    Kamba & jeova & jehova & 4K \\
    Tswana & basupi ba ga jehofa & jehova's witnesses & 4K \\
    Pedi & dihlatse tša jehofa & jehova's witnesses & 3K \\
    \midrule
    \rowcolor{gray!20} 
    \multicolumn{4}{c}{Betting } \\
    \midrule
    Indonesian & slot online & - & 33M \\
    Indonesian & judi online & online gambling & 24M \\
    North Azerbaijani & mostbet &  - & 20M \\
    Dutch & online casino & - & 13M \\
    Slovene & igralni avtomat & slot machine & 1M \\
    Khmer & jackpot party slots & - & 435K \\
    North Azerbaijani & pin up casino online & - & 265K \\
    North Azerbaijani & up on line casino & - & 200K \\
    Sundanese & online games & - & 111K \\
    Javanese & tohan maén bal & soccer betting & 35K \\
    Javanese & totoan piala donya & world cup betting & 17K \\
    \midrule
    \rowcolor{gray!20} 
    \multicolumn{4}{c}{Adult Content} \\
    \midrule
    Spanish & prostitutas & prostitutes & 549M \\
    French & site de rencontre & dating site & 100M \\
    Norwegian Bokmål & sex & - & 67M \\
    French & plan cul & sex date & 57M \\
    Norwegian Bokmål & dating & - & 56M \\
    Finnish & porno & porn & 55M \\
    Finnish & sex & - & 40M \\
    Swedish & video porno & porn video & 23M \\
    Swedish & erotisk massage & erotic massage & 16M \\
    Italian & donna cerca uomo & woman seeking man & 13M \\
    Dutch & erotische massage & erotic massage & 13M \\
    Swedish & escort tjejer & scort girls & 10M \\
    Finnish & eroottinen hieronta & erotic massage & 8M \\
    Icelandic & kynlíf & sex & 7M \\
    Icelandic & vændiskonur & prostitutes & 7M \\
    Danish & sex massage & - & 7M \\
    Icelandic & klám & porn & 6M \\
    Danish & body to body massage & - & 925K \\

    \bottomrule
    \end{tabular}}
    \caption{Frequent \textit{n}-grams in monolingual datasets.}
    \label{tab:mono_ngrams}
\end{table*}

\begin{table*}[!h]
\centering\small
\setlength{\tabcolsep}{1.2ex}
\adjustbox{max width=\textwidth}{%
\begin{tabular}{lrrrrr}
\toprule
\textbf{} & \multicolumn{1}{c}{\textbf{Raw}} & \multicolumn{1}{c}{\textbf{Filtered}} &\multicolumn{3}{c}{\textbf{TMX}} \\
\cmidrule(lr){2-2} \cmidrule(lr){3-3} \cmidrule(lr){4-6}
\textbf{Language} & \textbf{Sentence Pairs} & \textbf{Sentence Pairs} & \textbf{Sentence Pairs} & \textbf{English Words} & \textbf{Words / Sentence}\\
\midrule
Norwegian Nyorsk & 1 392 393 & 347 630 & 199 114 & 2 945 773 & 14.79\\
Georgian & 2 869 087 & 1 942 455 & 1 500 554 & 37 227 421 & 24.81\\
Maltese & 10 254 459 & 3 668 752 & 2 434 798 & 59 538 368 & 24.45\\
Basque & 7 014 653 & 3 848 389 & 2 614 143 & 53 581 439 & 20.50\\
Galician & 7 887 381 & 5 163 954 & 3 548 466 & 69 533 353 & 19.60\\
Irish & 9 526 442 & 5 633 190 & 3 916 880 & 77 433 464 & 19.77\\
Icelandic & 20 116 835 & 11 583 226 & 5 491 654 & 96 921 088 & 17.65\\
Macedonian & 11 456 233 & 8 077 748 & 5 818 642 & 111 266 193 & 19.12\\
Bosnian & 31 536 332 & 10 567 604 & 6 822 086 & 123 946 341 & 18.17\\
Albanian & 13 887 717 & 9 313 105 & 6 864 132 & 139 971 829 & 20.39\\
Serbian & 20 220 519 & 11 946 813 & 8 495 542 & 163 296 465 & 19.22\\
Estonian & 41 113 567 & 24 100 239 & 15 520 792 & 315 685 457 & 20.34\\
Slovenian & 41 381 480 & 25 825 392 & 16 845 959 & 347 356 986 & 20.62\\
Catalan & 44 286 113 & 29 337 960 & 17 601 213 & 366 406 455 & 20.82\\
Latvian & 52 450 307 & 32 891 012 & 18 743 189 & 370 988 789 & 19.79\\
Lithuanian & 60 997 986 & 37 873 083 & 21 328 192 & 417 168 369 & 19.56\\
Croatian & 76 512 486 & 45 523 434 & 26 377 878 & 508 387 461 & 19.27\\
Slovak & 84 379 108 & 56 064 987 & 32 237 055 & 614 417 180 & 19.06\\
Norwegian Bokmål & 87 684 923 & 57 244 547 & 34 528 565 & 622 620 428 & 18.03\\
Bulgarian & 84 816 588 & 56 252 074 & 35 660 392 & 680 913 110 & 19.09\\
Turkish & 117 558 747 & 67 850 467 & 38 800 144 & 858 856 570 & 22.14\\
Ukranian & 95 060 018 & 67 435 962 & 45 402 157 & 851 894 055 & 18.76\\
Hungarian & 130 875 234 & 79 142 830 & 45 500 610 & 880 509 225 & 19.35\\
Finnish & 132 307 703 & 84 100 057 & 47 622 955 & 869 456 254 & 18.26\\
Romanian & 140 205 353 & 85 394 851 & 54 481 584 & 1 067 957 734 & 19.60\\
Greek & 151 898 543 & 98 232 054 & 55 649 558 & 1 058 129 894 & 19.01\\
Czech & 153 423 740 & 99 670 275 & 57 873 634 & 1 091 697 726 & 18.86\\
Danish & 185 119 314 & 121 732 584 & 68 815 563 & 1 256 911 174 & 18.26\\
\midrule
Total & 1 816 233 261 & 1 140 764 674 & 680 695 451 & 13 115 018 601 & 19.62\\
\bottomrule
    \end{tabular}}    
    \caption{Statistics for the parallel portion of HPLT 3.0 before filtering (Raw), after Bicleaner AI (Filtered) and after
deduplication (TMX). Languages are in increasing order of deduplicated sentence pairs.}
    \label{tab:bitext-stats}
\end{table*}

\begin{table*}[!h]
\centering\small
\setlength{\tabcolsep}{1.2ex}
\begin{tabular}{lrrrrrr}
\toprule
 & \multicolumn{3}{c}{\textbf{XX - English}} & \multicolumn{3}{c}{\textbf{English - XX}} \\
 \cmidrule(lr){2-4} \cmidrule(lr){5-7}
\textbf{Language} & \textbf{BLEU} & \textbf{ChrF} & \textbf{COMET} & \textbf{BLEU} & \textbf{ChrF} & \textbf{COMET}\\
\midrule
Albanian & 35.19 & 62.95 & 82.15 & 30.68 & 59.48 & 82.38\\
Basque & 26.19 & 55.37 & 79.96 & 18.55 & 57.04 & 78.77\\
Bosnian & 37.59 & 64.40 & 82.19 & 31.59 & 60.74 & 82.94\\
Bulgarian & 37.61 & 65.03 & 81.66 & 41.10 & 67.36 & 84.16\\
Catalan & 42.02 & 67.69 & 81.69 & 41.34 & 65.99 & 80.46\\
Croatian & 34.46 & 61.82 & 80.95 & 30.67 & 59.82 & 82.58\\
Czech & 36.29 & 63.62 & 81.15 & 31.71 & 58.77 & 81.37\\
Danish & 44.06 & 69.13 & 84.85 & 45.53 & 69.14 & 83.69\\
Estonian & 34.22 & 61.65 & 82.86 & 27.02 & 60.34 & 84.16\\
Finnish & 29.52 & 58.00 & 82.61 & 23.65 & 58.28 & 85.52\\
Galician & 35.21 & 63.37 & 81.42 & 32.92 & 60.32 & 79.47\\
Georgian & 20.65 & 50.44 & 75.88 & 12.98 & 51.17 & 75.97\\
Greek & 30.88 & 59.19 & 81.27 & 27.07 & 53.59 & 83.17\\
Hungarian & 31.72 & 60.23 & 81.33 & 27.40 & 58.42 & 81.65\\
Icelandic & 31.01 & 57.36 & 79.74 & 26.03 & 53.82 & 76.86\\
Irish & 36.47 & 63.06 & 77.54 & 33.77 & 60.43 & 73.25\\
Latvian & 33.03 & 61.53 & 81.57 & 31.89 & 60.50 & 82.11\\
Lithuanian & 30.15 & 57.84 & 79.39 & 27.75 & 58.94 & 82.34\\
Macedonian & 38.40 & 64.86 & 81.66 & 34.79 & 63.71 & 81.75\\
Maltese & 49.20 & 72.66 & 76.79 & 39.08 & 70.69 & 69.76\\
Norwegian Bokmål & 39.06 & 65.20 & 82.84 & 33.70 & 62.10 & 83.77\\
Norwegian Nyorsk & 20.27 & 45.00 & 62.40 & 9.57 & 33.88 & 54.64\\
Romanian & 39.53 & 65.75 & 83.47 & 39.37 & 64.04 & 83.25\\
Serbian & 39.43 & 65.98 & 80.96 & 35.23 & 62.51 & 81.37\\
Slovak & 34.17 & 62.41 & 80.61 & 33.30 & 60.29 & 81.44\\
Slovenian & 31.72 & 59.89 & 81.38 & 30.10 & 58.00 & 81.43\\
Turkish & 34.22 & 61.07 & 82.16 & 28.56 & 60.43 & 80.14\\
Ukranian & 35.53 & 62.43 & 80.45 & 28.72 & 57.61 & 81.51\\
\bottomrule
    \end{tabular}
    \caption{MT results (BLEU, Chrf, COMET) for models translating the FLORES200 devtest benchmark from English and into English trained on our HPLT 3.0 dataset.}
    \label{tab:mt-models}
\end{table*}

\end{document}